\documentclass[sn-mathphys-num, iicol, pdflatex]{sn-jnl}

\usepackage{graphicx}%
\usepackage{multirow}%
\usepackage{amsmath,amssymb,amsfonts}%
\usepackage{amsthm}%
\usepackage[title]{appendix}%
\usepackage{xcolor}%
\usepackage{textcomp}%
\usepackage{manyfoot}%
\usepackage{booktabs}%
\usepackage{listings}%

\usepackage[ruled, lined, longend, linesnumbered]{algorithm2e}
\usepackage{enumitem}
\usepackage{rotating}

\theoremstyle{thmstyleone}%
\theoremstyle{thmstyletwo}%

\theoremstyle{thmstylethree}%

\begin{document}

\title[Robots Influencing Humans to Reveal their Goals during Collaboration and Competition]{Robots Influencing Humans to Reveal their Goals during Collaboration and Competition}


\author*[1]{\fnm{Debasmita} \sur{Ghose}}\email{debasmita.ghose@yale.edu}
\equalcont{These authors contributed equally to this work.}

\author*[1]{\fnm{Oz} \sur{Gitelson}}\email{oz.gitelson@yale.edu}
\equalcont{These authors contributed equally to this work.}

\author[2]{\fnm{Michal} \sur{Lewkowicz}}\email{michal01@mit.edu}

\author[3]{\fnm{Jake} \sur{Brawer}}\email{jake.brawer@colorado.edu}

\author[3]{\fnm{Alessandro} \sur{Roncone}}\email{alessandro.roncone@colorado.edu}

\author[1]{\fnm{Marynel} \sur{Vazquez}}\email{marynel.vazquez@yale.edu}

\author[1]{\fnm{Brian} \sur{Scassellati}}\email{brian.scassellati@yale.edu}

\affil[1]{\orgdiv{Department of Computer Science}, \orgname{Yale University},
\orgaddress{\street{51 Prospect Street}, 
\city{New Haven}, \postcode{06511}, \state{CT}, \country{USA}}}

\affil[2]{\orgdiv{Department of Aeronautics and Astronautics}, \orgname{Massachusetts Institute of Technology}, 
\orgaddress{\street{125 Massachusetts Ave}, \city{Cambridge}, \postcode{02139}, \state{MA}, \country{USA}}}

\affil[3]{\orgdiv{Department of Computer Science}, \orgname{University of Colorado, Boulder}, 
\orgaddress{\street{430 UCB, 1111 Engineering Dr}, \city{Boulder}, \postcode{80309}, \state{CO}, \country{USA}}}


\abstract{We propose a unified strategy for fast goal inference in human–robot interaction. The core idea is to drive the human toward \textit{Critical Decision Points (CDPs)}—states where competing human strategies prescribe different next actions and thus maximally reveal the goal. We formalise CDPs using a goal‑conditioned policy divergence measure and incorporate them into a Receding‑Horizon Planner that explores future action sequences while optimizing a cost function balancing task progress and information gain. We evaluate this approach in both a collaborative, fully observable cooking task and a competitive, partially observable hide‑and‑seek game, each in simulation and on real robots. In both scenarios, our method infers human goals more accurately and earlier than baseline strategies.}


\maketitle

\section{Introduction}

Robots that interact with people in shared environments must understand the high-level intentions or goals that guide human behavior \cite{hoffman2024inferring}. In both competitive and collaborative situations, a robot's ability to anticipate what a person is trying to do or what their goal is, rather than merely reacting to their previous actions, enhances the efficiency of its actions and supports more natural interactions \cite{dhakal2024anticipatory}. However, goal inference is challenging: the shared environment can be partially observable, sensing is often noisy, and many human high-level goals often share initial actions. Standard approaches that passively observe the human and update a Bayesian belief assume frequent, distinctive observations or fully observable worlds. Under sparse, ambiguous, or adversarial conditions, these methods converge too slowly, leaving the robot uncertain for a significant portion of the interaction.

To allow a robot to quickly understand a human's goal early in the interaction, we have identified a key insight: \textit{certain states provide much more informative cues about people's objectives than others.} We call such states \textit{Critical Decision Points (CDPs)}. A CDP is any state in the shared environment of the human and the robot where two or more plausible human strategies would prescribe different immediate actions, making the human’s next move highly revealing to a robot. Intuitively, at CDPs, subsequent actions of the human would clearly differentiate between competing goals. 

We hypothesize that by steering the interaction toward CDPs, the robot can reduce uncertainty more rapidly than if it were to rely solely on passive observation. Thus, we propose that the robot use Receding Horizon Planning (RHP) \cite{ma2006receding} to actively influence the human toward CDPs to reveal their goal earlier in the interaction than passive observation. 


In this work, we show that the insight of identifying CDPs and influencing the human towards those CDPs is valuable in both competitive and collaborative tasks, in fully or partially observable environments, and remains effective whether the actions taken by the human and the robot are temporally independent or unfold over long action histories.
We validate this idea in two scenarios: a collaborative task performed in a fully observable environment with strongly time-dependent actions, and a competitive task executed in a partially observable setting where actions are largely temporally independent.

In many real-world tasks, multiple overlapping action sequences can be used to accomplish different goals. For example, the steps for assembling a chair are similar to those for a stool. In meal preparation, making pancakes involves actions like mixing batter and pouring it onto a pan, which closely resemble those needed for muffins. This overlap can make it challenging for a robot collaborating with a human to determine a human's specific goal, and make goal inference via passive observation unreliable. 
Moreover, traditional goal inference in human-robot collaboration has relied heavily on communication between humans and robots \cite{kaupp2010human, chang2018effects, nikolaidis2013human, unhelkar2020decision, st2015robot, lemasurier2021methods, 10610720, brawer2018situated, chen2024human, ghose2026open, macke2021expected, ghose2026job}. However, verbal communication between agents may not always be feasible due to various physical or situational constraints \cite{ghose2023tailoring, hulle2024eyes, liu2018goal}. Even when verbal communication is possible, the task may be too complex to explain clearly, or humans and robots might interpret the same task description differently \cite{agrawal2022task, bobu2024aligning, ghose2025adapting}. {\color{black}For example, if a person tells a robot, “I want oatmeal for breakfast,” this still leaves many preferences about how the human wants their oatmeal unclear, such as whether to use milk or water, which toppings to prepare, and in what order to carry out the steps. A robot can resolve these ambiguities by asking many follow-up questions or by requiring the person to specify the task in step-by-step detail, but both options increase the burden on the human.} Disabilities or the impracticality of verbal communication in specific contexts, such as when it might interrupt another conversation or concurrent tasks the human might be doing, can also make it challenging for humans to convey their intentions directly \cite{mannem2023exploring}.

To address the challenges of overlapping action sequences and the absence of verbal communication, we propose that the robot guide the human toward CDPs which are states that could reveal their goal early in the collaboration. To guide the human towards CDPs, the robot expands a Receeding Horizon Planning (RHP) tree, to imagine many possible reasonable future actions of the human and the robot, and selects an action that maximizes the amount of information gained while balancing task performance. 

We evaluate our approach to influence a human toward CDPs in a collaborative cooking setup, both in simulation and with a physical robot, where a human and a robot prepare up to 30 meals collaboratively. 
{\color{black}We chose cooking because it provides concrete ground truth for controlled experiments, supports comparison with prior work \cite{brawer2023interactive, carroll2019utility, van2022correct, goubard2023cooking, ghose2025ve, ghose2026open}, and includes many high-level goals that share initial action sequences.} 
Results show that clarifying human intentions sooner enables the robot to adjust its actions more effectively, aligning them better with the human's goal and reducing redundant or conflicting efforts.  We compare our approach with prior works that have addressed the challenge of goal inference using various strategies. Information gain maximization methods, such as \cite{sadigh2016information}, focus on influencing the human towards taking actions that increase the robot's information gain without fully considering whether the robot’s actions align with the human’s true goal, leading the robot to take many incorrect actions during collaboration, especially when goals have overlapping action sequences. Bayesian Delegation \cite{wu2021too}, on the other hand, continuously updates its belief over possible goals based on observed actions. However, the method tends to overemphasize these overlapping action sequences, diluting the belief in possible goals and making it harder for the robot to disambiguate between them confidently. 

{\color{black}Next, we demonstrate our approach in a partially-observable setting through a hide-and-seek game between a human and a robot (Fig.~\ref{fig:realrobot}). In this game, the seeker robot tries to catch a human-controlled hider by estimating their strategy.
Since the environment can be discretized into a grid, we demonstrate how CDPs can be precomputed over the spatial state space. }

Our experiments in simulation and our demonstration in the real world show that when the robot seeker influences the hider toward CDPs, the seeker can catch the hider more quickly than when the robot attempts to estimate the human's strategy while randomly exploring the map or minimizing the distance between itself and the hider whenever the hider is visible. Overall, our findings suggest that robots can leverage the properties of a shared environment in interactive human-robot tasks to estimate a human's strategy under partial observability. 

In summary, our contributions are:
\begin{itemize}[leftmargin=*]
\item  We show that human intentions can be revealed much earlier if the robot \emph{actively drives the interaction toward a small set of highly informative states}. We call these states Critical Decision Points (CDPs), states where competing human strategies prescribe different immediate actions.
\item  We provide a formal, goal‑conditioned definition of CDPs and an accompanying Receding‑Horizon planner that steers the human toward them while trading off task progress and information gain.
\item {\color{black}We evaluate the same CDP-based goal inference framework in two settings with contrasting interaction assumptions: a collaborative, fully observable cooking task with time-dependent actions, and a competitive, partially observable hide-and-seek task with largely independent actions. To our knowledge, this is the first unified goal-prediction approach shown to operate effectively across both cooperative and competitive interactions, yielding earlier and more accurate inference than baselines.}

\end{itemize}
\section{Related Work}

\subsection{Robots Predicting Human Goals}

\subsubsection{Goal Prediction in Human-Robot Collaboration}

Inferring human goals and predicting their actions is crucial for effective human-robot collaboration, enabling robots to provide assistance \cite{hoffman2024inferring}. Researchers have studied this in contexts such as predicting handover goal positions \cite{zhuang2022goferbot, choi2022preemptive, laplaza2022context, ghose2025adapting}, reasoning over human plans in collaborative workspaces \cite{hoffman2007cost, nikolaidis2013human, haninger2022model, el2022hierarchical, adamson2021we, tung2024workspace}, inferring intentions in shared autonomy \cite{jain2019probabilistic, jonnavittula2021know, qiao2021learning, aronson2021inferring, aronson2022gaze, decastrodreaming}, and predicting motion for social navigation \cite{thompson2009probabilistic, bandyopadhyay2013intention, kollmitz2015time, thompson2024predicting}.

Several key approaches have been employed to infer human goals and intentions in human-robot collaboration. These include direct Bayesian inference, which calculates the posterior probability of each possible goal, either in a single step \cite{pellegrinelli2016human, jain2019probabilistic, iregui2021reconfigurable, jonnavittula2021know, felip2022intuitive} or through dynamic processes like Markov models or MDPs \cite{hoffman2007cost, cramer2021probabilistic, zhao2022coordination, losey2022physical, bandyopadhyay2013intention, koppula2016anticipatory, hong2007mixed}. Another group of methods uses supervised learning techniques, such as neural networks, to map observations to probabilities over intentions using training data \cite{schrum2022mind, haninger2022model, urkmez2022detecting, cui2023no, qin2024learning}. {\color{black}Freedman et al. \cite{freedman2019responsive} propose an approach for the robot to execute a weighted combination of precomputed policies for different goals based on its belief.} These methods generally take a passive approach to human goal inference, relying heavily on the observation of human actions to deduce intent after the actions have been performed.

Researchers have also explored more active strategies to reveal human goals. 
{\color{black}One line of work casts the robot as a passive observer that intervenes solely to disambiguate human goals, without contributing to task execution \cite{amato2019active, shvo2020active}}. Another approach is Bayesian Delegation \cite{wu2021too}, which uses Bayesian inference to predict a human's response to a robot's action, enabling effective human-robot collaboration. Prior work has also demonstrated that robots can take information-gathering actions to better understand human internal states for autonomous driving tasks using Model Predictive Control \cite{sadigh2016information}, also without contributing to shared task completion. 
In this work, we propose Critical Decision Points to enable the robot to quickly infer human goals during a sequential human-robot collaboration task in an online manner, particularly for tasks where overlapping action sequences result in multiple goals. We compare our method with 
an Information Gain Maximization approach \cite{sadigh2016information} and  Bayesian Delegation \cite{wu2021too} for a collaborative cooking task, mimicking real-world cooking tasks where multiple goal recipes could contain many overlapping action sequences, and show that our method is more effective at inferring human goals earlier during the task. 

\subsubsection{Opponent Modeling}

In competitive settings, opponent modeling is used to build a model of an opponent’s behavior using observations of their actions  \cite{nashed2022survey, tian2023multi, yu2022model}. Over the years, opponent modeling has been extensively studied for applications such as playing stochastic games between agents \cite{von2017minds}, negotiations between agents \cite{nazari2015opponent}, and for cooperative tasks \cite{wang2021emergent, yang2020learning}. For human-robot interaction tasks, prior work uses Theory of Mind  \cite{oguntola2023theory, rabinowitz2018machine, von2017minds} to recursively reason about the human's actions based on the robot's behavior. Some methods use inverse reinforcement learning \cite{abbeel2004apprenticeship} to 
estimate a human’s reward model \cite{bonjour2022decision, bui2023imitating} and then use the reward model to estimate optimal robot behavior via reinforcement learning techniques. 
{\color{black}Bisson et al. \cite{bisson2011provoking} treat the robot as a reactive observer in competitive games, using handcrafted provocative events to disambiguate the opponent’s goal without actively engaging in the task. Therefore, we build on these ideas by enabling the robot to participate in the game actively: our robot uses a Bayesian approach to continuously estimate the human’s strategy while steering the interaction toward Critical Decision Points—states where human behavior is most revealing of their underlying goal.}

\subsection{Robots Influencing Humans}

In recent years, there has been a growing interest in enabling robots to select actions that influence human behaviors \cite{pmlr-v164-wang22f, ghose2024planning, tian2023towards, li2021influencing, valtazanos2013bayesian, sagheb2023towards, nikolaidis2017human, pandya2024towards, chari2024optimal, bisson2011provoking} in both collaborative and competitive scenarios.  A common approach is to explicitly model the human’s reward \cite{abbeel2004apprenticeship} and then optimize for robot actions that, at least in the near future, steer the human towards states where they are more likely to act in a desired manner \cite{sadigh2016planning}. More recently, high-level strategies of humans have been inferred from repeated interactions using unsupervised learning \cite{parekh2023learning, xie2021learning} or theory-of-mind approaches \cite{favier2022robust, shvo2023theory, tsoi2026coadaptive}, with the inferred strategies being used to influence their behavior. Prior work has also leveraged offline reinforcement learning to train robots to influence humans more optimally \cite{hong2024learning, pmlr-v164-wang22f, pandya2024robots}. Building upon this line of research, we frame the problem of influencing humans  to reveal their true goal as a Receding Horizon Planning problem \cite{ma2006receding, renz2024moving}. We do this by enabling both competitive and collaborative robots to select actions that drive a human towards states where their actions are more revealing about their goal.

\section{Influencing Humans to Reveal their Goals at Critical Decision Points}

\subsection{Preliminaries}

\subsubsection{State Space}
We consider a robot working alongside a person in a shared environment and aim to enable the robot to infer the person’s high-level strategy or goal. We define the full environment state to be
$s = (\,s_w,\;s_h,\;s_r\,),$
where $s_w$ is a discrete description of the world (for example, an occupancy grid or the locations of key objects), $s_h$ is the human’s state {\color{black}(e.g., position or orientation)}, $s_r$ is the robot’s state. {\color{black}The robot does not directly observe $s_h$; instead, it receives observations $o_h$ through its sensors (for example, an RGB-D camera or lidar), which may be noisy or partial depending on the task context.} If the environment were fully observable, we assume the robot could have access to the full history $\mathcal{H}$ of all the actions taken during the interaction by the human. In a partially observable environment, since $s_h$ is not always complete, we cannot assume access to the full history $\mathcal{H}$. 
{\color{black}Additionally, to balance participation between the human and the robot, we assume the interaction is turn-based.}
{\color{black}We assume that the robot plans over a task-level representation of the shared environment. This representation contains the information needed to determine which actions the human and robot can take from the current state and what states those actions may lead to. The representation may be symbolic, discrete, or a discretized abstraction of the physical world.}

\subsubsection{Goal and Policy Bank}
We assume a finite \textit{goal set}
$\mathcal{G} = \{\,g_1,\dots,g_n\},$ where each $g$ denotes a distinct high‑level goal the human might pursue (for example, a particular destination in a navigation task or a specific dish in a cooking task). {\color{black}The robot is given this candidate goal set for the interaction, but does not know which goal the human is pursuing.}  To each goal $g$ we associate a distribution of policies $\mathcal P_g$, where each policy in $\mathcal P_g$ is possibly stochastic and captures ``how the human would act if their true objective were $g$”.  Concretely, assume that $\mathcal P_g$ has $n$ policies. Each policy $i \in [1,n]$ in $\mathcal P_g$, $\pi_g^i$, defines the human’s next action $a_h \in\mathcal A_h$, conditioned on the current state $s$ and, when relevant, the interaction history $H$, where $\mathcal A_h$ signifies all possible human actions. 
Formally,
\begin{align}
     \pi_g^i(a_h\mid s,H)
  = P\bigl(a_h \mid s, H,\text{goal}=g\bigr).
\end{align}
 
In the remainder of this paper, we refer to $\mathcal P_g$ as the \textit{policy bank} corresponding to goal $g$.
For example, in competitive navigation, the policy bank $\mathcal P_{\text{corner}}$ might include several different evasion paths all ending at the same goal location $corner$. In collaborative cooking, $\mathcal P_{\text{omelette}}$ could encompass different valid ingredient‑ordering strategies for making an $omelette$. 
{\color{black}Intuitively, each policy in the bank defines a mapping from state (and optionally history) to an action distribution. The policy bank includes realistic ways a human might act to achieve the goal in the environment.}
Finally, we denote \textit{full policy bank} as the union of policies for every goal $g \in \mathcal{G}$, $\displaystyle \mathcal P = \bigcup_{g\in\mathcal G} \mathcal P_g$.  We also assume the human’s actual goal‑policy $\pi_g^*$ remains constant throughout each interaction (i.e.\ they do not switch goals mid‐task), {\color{black} although the specific actions taken to pursue that goal may vary according to the policies in the corresponding policy bank.}

\subsubsection{Belief over Goals}
\label{sec: belief}

At each time step, the robot maintains a belief  
\begin{equation}
     b_{t-1}(g)\;=\;
  P\!\bigl(\text{goal}=g \mid \mathcal H_{t-1}\bigr),
\end{equation}

\noindent
where \(\mathcal H_{t-1}\) is the sequence of observed human actions up to
time \(t-1\).  
After the robot observes the human perform action \(a_h^{t}\) in state
\(s_{t}\), it updates this belief using a Bayes filter.  
{\color{black}We assume a uniform prior over goals, meaning all goals are equally likely before any actions are observed. This prior is folded into the belief $b_{t-1}(g)$, so subsequent updates do not explicitly show a prior term.}

Each goal $g$ can be accomplished not by a single policy but by executing a policy from a \emph{policy
bank}  
$\mathcal P_g=\{\pi_g^{1},\dots,\pi_g^{n_g}\}$
Hence, the likelihood of an action must marginalise over all policies in that
bank:  
\begin{equation}
  \bar{\pi}_{g}(a \mid s,H)
  \;=\;
  \frac{1}{|\mathcal P_g|}
  \sum_{i=1}^{n_g}
     \pi_g^{i}(a \mid s,H).
  \label{eq:marginal_pi}
\end{equation}
Intuitively, $\bar{\pi}_{g}$ answers the question:  
\emph{“If the human’s goal were \(g\) and they could be following any valid
policy in \(\mathcal P_g\) with equal prior weight, how likely is the action
I just saw?”}  
The belief update is therefore  
\begin{equation}
  b_{t}(g)=
  \frac{
        \bar{\pi}_{g}\!\bigl(a_h^{t}\mid s_{t},\mathcal H_{t-1}\bigr)\,
        b_{t-1}(g)}
       {\displaystyle
        \sum_{g'}
        \bar{\pi}_{g'}\!\bigl(a_h^{t}\mid s_{t},\mathcal H_{t-1}\bigr)\,
        b_{t-1}(g')}.
  \label{eq:bayes_goal_update}
\end{equation}

{\color{black}We can also reinterpret the Bayes update in terms of observed state transitions rather than actions.
In this view, the robot asks: \textit{``What is the likelihood of reaching state $s_t$, if the human were pursuing goal $g$?"}. If the transition model were deterministic
\(T(s_{t}\mid s_{t-1},a_h^{t})=1\), $P(s_{t}\mid g)$ would be computed as:}

\begin{align}
\label{eq:belief}
     P(s_{t}\mid g) &=
  \sum_{a} T(s_{t}\mid s_{t-1},a)\,
               \bar{\pi}_{g}\bigl(a\mid s_{t-1},\mathcal H_{t-1}\bigr) \notag \\ 
  &=\bar{\pi}_{g}\bigl(a_h^{t}\mid s_{t-1},\mathcal H_{t-1}\bigr),
\end{align}
so the standard Bayes filter gives  
\begin{equation}
  b_{t}(g)\;\propto\;
  P(s_{t}\mid g)\,b_{t-1}(g).
  \label{eq:belief_final}
\end{equation}

Here $b_{t-1}(g)$ is the prior at time $t-1$, and the likelihood
$\bar{\pi}_{g}$ (or equivalently $P(s_{t}\mid g)$ is estimated
empirically from state–visitation frequencies generated by all policies in
$\mathcal P_g$. {\color{black} Note that, in Eq. \ref{eq:belief}, we show the update under a deterministic transition function (\(T(s_{t}\mid s_{t-1},a_h^{t})=1\)), where a given action leads to a unique next state. If the transition function is stochastic, the same update can be used by computing the likelihood of the observed next state under each goal-conditioned policy.}

\begin{figure}
    \centering
    \includegraphics[width=\linewidth]{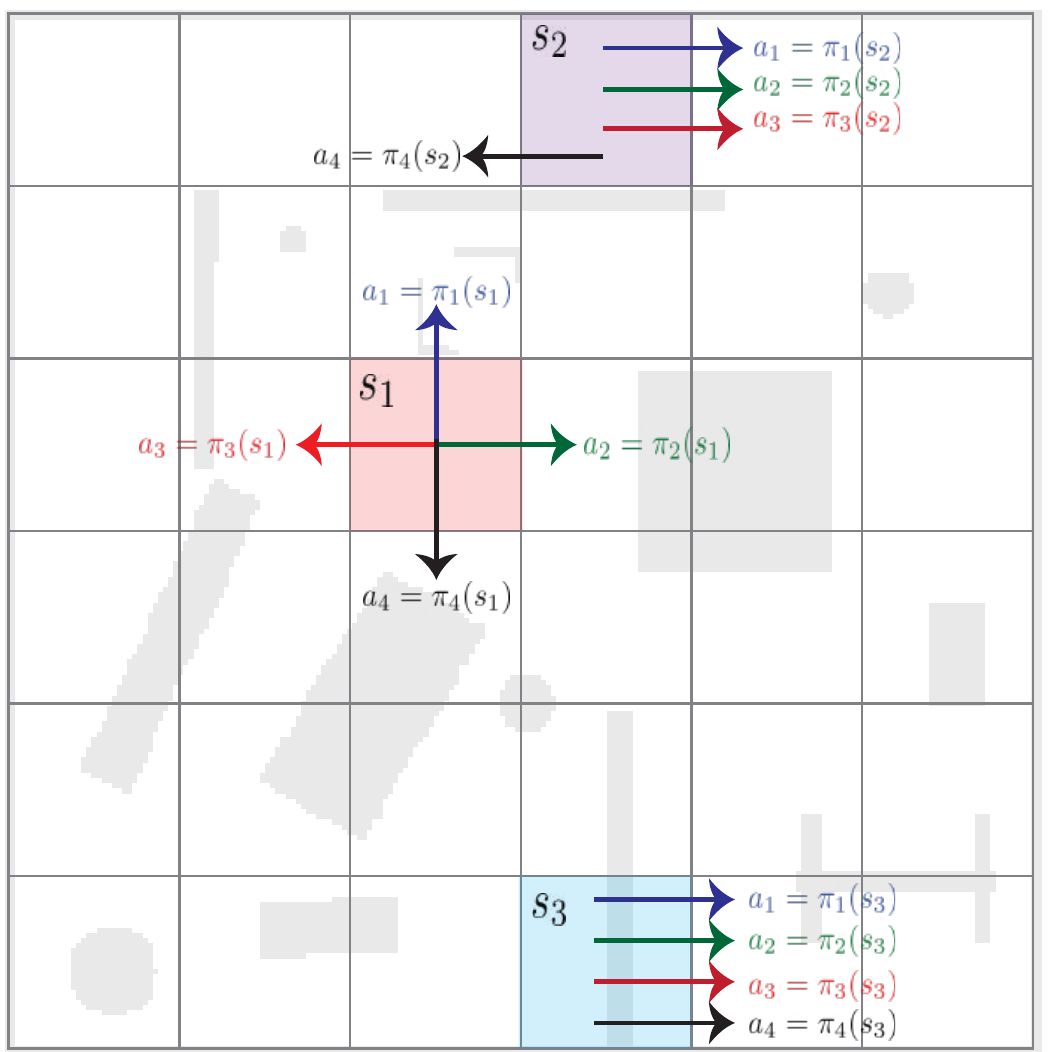}
    \caption{\textbf{Critical Decision Point Computation}. {\color{black} For each state, the robot computes how much the action distributions induced by policies for different candidate goals disagree. States with high disagreement are Critical Decision Points because observing the human’s next action helps distinguish between goals. }In the example, $s_1$ is a CDP because the policies associated with different goals prescribe different actions, while $s_2$ and $s_3$ are less informative compared to $s_1$ because the predicted actions largely overlap.}
    \vspace{-1em}
    \label{fig:CDP}
\end{figure}

\subsection{Critical Decision Points}
\label{sec:cdp}

The primary insight of our work is that certain states in the environment are inherently more revealing of the human’s underlying goal than others.  We call these states \textit{Critical Decision Points (CDPs)}.  Different goal‐conditioned policies prescribe markedly different next actions at a CDP, so observing which action the human actually takes yields maximal information about their true objective.

Mathematically, suppose our goal set is $\mathcal G=\{g_1,\dots,g_n\}$ and each goal $g_i$ is associated with a policy bank $P_{g}$.  At any state $s$, and given interaction history $H$ if available (in fully observable environments), each goal is represented by the marginal action distribution $\bar{\pi_g^i}$ as described in eq. \ref{eq:marginal_pi} over the human's next action. We measure the total pairwise disagreement among these distributions by

\begin{align}
\label{eq: cdp}
  M(s,H)
  = \sum_{1 \le i < j \le n}
      D\!\bigl(\bar{\pi}_{g}^i(a_h\mid s,H),\,\bar{\pi}_{g}^j(a_h\mid s,H)\bigr),
\end{align}

\noindent
where $D(p,q)$ is any symmetric divergence between two probability distributions (for example, the total‐variation or $\ell_1$ distance or a raw count of the number of policies diverging at a state). 
{\color{black}This total, $M(s,H)$, captures the extent of disagreement among goal-conditioned policies to identify states where multiple policies diverge meaningfully.} {\color{black}In our setting, this disagreement is interpreted with respect to the goals associated with the policies. That is, a state is informative when policies associated with different candidate goals prescribe different actions, so observing the human’s next action helps distinguish between those goals. Differences between two policies that lead to the same goal are not sufficient by themselves to make a state a CDP unless they also help separate that goal from other candidate goals.}
A state $s$ becomes a CDP when this disagreement between policies {\color{black}associated with different candidate goals} exceeds a chosen threshold $\theta$:

\begin{align}
   S_{\rm CDP}(H)
     = \bigl\{\,s \in \mathcal S \;\bigm|\;
       M(s,H) \;>\;\theta
     \bigr\},
\end{align}

{\color{black}Intuitively, in Fig.~\ref{fig:CDP}, there are four policies, each leading to a different goal. At state $s_1$, each policy selects a distinct action—up, right, left, and down—so the pairwise divergence $M(s_1, H)$ would be 4, indicating high disagreement and making $s_1$ a CDP. In contrast, at $s_3$, all policies choose the same action (right), yielding a divergence of 0 and making $s_3$ uninformative for inferring the human’s true goal.}


CDPs can be precomputed offline when no access to history is available by evaluating $M(s,\varnothing)$ once for every state and storing those above the threshold, and can also be computed online during interaction by recalculating $M(s,H)$ with the current history.

Our central hypothesis is that by steering the human and the shared interaction toward these CDPs, the robot can observe the most distinguishing human actions and thus reduce its uncertainty over the human’s goal far more quickly than by passive observation alone.

\begin{algorithm}[t]
\small
\caption{\emph{Overview} - Influencing the Human toward CDPs with RHP}
\label{alg:overview_cdp}
\KwIn{Goal set $\mathcal G=\{g_1,\dots,g_n\}$; policy banks
       $\mathcal P_g=\{\pi_g^{1},\dots,\pi_g^{n_g}\}$;
       horizon $D_H$; weights $\lambda_{1,2}$;
       divergence $D$; task cost $C_{\!\mathrm{task}}$}
\KwOut{Next robot action $a_r^{\,t}$}

\textbf{Init:}\;
$b_{0}(g)\leftarrow1/|\mathcal G|$, \,
$\mathcal H\!\leftarrow\!\varnothing$, \,
$t\!\leftarrow\!0$\;

\While{task not finished}{
  \tcp{1) Sense}
  observe $s_t=(s_w,s_h,s_r)$\;

  \tcp{2) Expand depth‑$D_H$ tree}
  Alternate levels of:
  \begin{itemize}[leftmargin=*,nosep]
    \item robot branches over $a_r\in\mathcal A_r$,
    \item human samples
      $a_h\!\sim\!\bar{\pi}_{\hat g}(a_h\mid s,H)$
      $\displaystyle
        \bar{\pi}_{g}(a\!\mid\!s,H)=\frac1{|\mathcal P_g|}
          \!\sum_k\!\pi_g^k(a\mid s,H)$. {\color{blue}(Eq. \ref{eq:marginal_pi})}
  \end{itemize}
  Store every leaf $\ell$ with $s_\ell$ and $\mathcal H_\ell$\;

  \tcp{3) Score leaves}
  \ForEach{leaf $\ell$}{
    \textbf{Informativeness Score:} $M_\ell\leftarrow\sum_{g^i<g^j}
       D\!\bigl(\bar{\pi}_{g}^i(a_h\mid s_\ell,\mathcal H_\ell),
                \bar{\pi}_{g}^j(a_h\mid s_\ell,\mathcal H_\ell)\bigr)$\{\color{blue}(Eq. \ref{eq: cdp})}
     \textbf{Task Score: }$C_\ell\leftarrow C_{\!\mathrm{task}}(s_\ell,\mathcal H_\ell)$\;
     $J_\ell\leftarrow\lambda_1C_\ell+\lambda_2M_\ell$\;
  }

  \tcp{4) Act}
  $\ell^\star\leftarrow\arg\min_\ell J_\ell$;\;
  $a_r^{\,t}\leftarrow$ first robot step on path $\ell^\star$;\;
  execute $a_r^{\,t}$\;

  \tcp{5) Update belief}
  observe $a_h^{\,t}$\;
  $
    b_{t+1}(g)=
      \dfrac{\bar{\pi}_{g}(a_h^{\,t}\mid s_t,\mathcal H)\,b_t(g)}
            {\sum_{g'}\bar{\pi}_{g'}(a_h^{\,t}\mid s_t,\mathcal H)\,b_t(g')}; {\color{blue}
            \text{(Eq. \ref{eq:bayes_goal_update})}}
  $\
  $\mathcal H\leftarrow\mathcal H\cup\{a_h^{\,t}\}$,\;
  $t\leftarrow t{+}1$\;
\end{algorithm}

\subsection{Influencing the Human to Critical Decision Points}
\label{sec:influencing}

We develop a method for the robot to reduce the uncertainty of its belief over the human's current goal, given the set of reasonable human goals in the goal bank $\mathcal{G}$, in both competitive and collaborative tasks. We frame the above problem as a discrete-time planning problem, which we solve using a Receding Horizon Planning (RHP) approach \cite{ma2006receding}. RHP takes inspiration from Model Predictive Control  \cite{garcia1989model} but concerns discrete decision-making rather than continuous control. Generally, to perform RHP, an agent iteratively solves a planning problem over a receding horizon, considering possible future actions by expanding a search tree. After planning on a given time step, the first action leading to the best possible outcome is executed, the horizon window moves forward, and planning is performed again for the next possible futures.

Algorithm \ref{alg:overview_cdp} summarizes our RHP-based approach to influence the human toward CDPs. 
At the start, the robot has no prior belief over the human’s objectives --- it initializes a uniform belief over all goals and an empty history of observed human actions.  The robot first perceives the current joint state $s_t=(s_w,s_h,s_r)$, where $s_w$ encodes the world state, $s_h$ the human’s (possibly noisy) state, and $s_r$ the robot’s own configuration.

Next, the robot builds a depth‑$D_H$ look-ahead tree of future states with the current state as the root. The tree consists of alternate robot and human moves. The robot action branches over every feasible action that can be taken from the current state. The potential human response to every robot action is drawn from the most likely goal‑conditioned policy $\bar{\pi}_g$ the robot believes the human is following.  

Each leaf $\ell$ is then given two scores: a \textit{task cost} $C_\ell$ that quantifies how well the robot’s path advances the human's goal, and an \textit{informativeness cost} $M_\ell$ that measures how strongly the goal hypotheses disagree on the human’s next move at $s_\ell$.  Crucially, this same $M$ is the divergence‑based measure used to define Critical Decision Points, states where different goals prescribe markedly different actions using Eq. \ref{eq: cdp}.  In other words, when $M_\ell$ is large, $s_\ell$ lies in the CDP set, and observing the human there would yield maximal clarity about their true goal.  The planner combines these into a single objective to favor futures that both progress the task and steer the human toward CDPs.
The robot selects the leaf with the lowest combined cost, executes the first robot move along that path, and waits for the human to respond. {\color{black}Thus, the robot is not selecting actions solely to infer the human’s goal; it is selecting the next robot action that advances its task objective while also making future human actions more informative. When reachable future states provide limited additional information, the informativeness term has less influence on the objective, and the selected action is driven more by task progress.}
Upon observing human actions after the robot has taken them in the real world, the robot refines its posterior over goals via Bayes’ rule and appends the new action to its history. Incrementing $t$, the tree is then expanded again to continually guide the interaction toward states that both support the task and most rapidly reduce uncertainty about the human’s intentions.

\section{Scenario 1: Influencing the Human to Reveal their Goal in Human-Robot Collaboration}



\begin{algorithm}
\small
\caption{Scenario 1: Selecting Action at Timestep t to Influence Humans to CDPs by expanding an RHP tree for Human-Robot Collaboration}
\label{algo}
\KwIn{%
  Goal set $\mathcal G=\{g_1,\dots,g_n\}$; per‑goal policy banks
  $\mathcal P_g=\{\pi_{g}^{1},\dots,\pi_{g}^{n_g}\}$;                                    \\
  Current joint history $\mathcal H^{0:t-1}$ of all observed actions;                    \\
  Planning horizon $D_H$; weights $(\lambda_1,\lambda_2)$;                              \\
  Divergence $D(\cdot,\cdot)$; task‑cost $C_{\!\mathrm{task}}(\cdot)$}
\KwOut{Next robot action $a_r^{\,t}$}

$b_{t}(g)\leftarrow 1/|\mathcal G|$ (no prior bias),\;
$\mathcal H\leftarrow \mathcal H^{0:t-1}$\;

\While{task not finished}{
  \tcp{1) Observe}
  sense $s_t=(s_w,s_h,s_r)$

  \tcp{2) Expand depth‑$D_H$ look‑ahead tree}
  Alternate robot branches ($a_r\in\mathcal A_r$) with human samples
  $a_h\sim \bar{\pi}_{\hat g}(a_h\mid s_t,\mathcal H)$,
  where 
  $\displaystyle
    \bar{\pi}_{g}(a\mid s,H)=\tfrac1{|\mathcal P_g|}\sum_k\pi_g^{k}(a\mid s,H)$. \hspace{13em} {\color{blue}(Eq. \ref{eq:marginal_pi})}\\
  Store each leaf $\ell$ with state $s_\ell$, augmented history $\mathcal H_\ell$
  (i.e.\ $\mathcal H$ extended with simulated actions), and future sequence
  $\mathcal F_\ell$.

  \tcp{3) Score leaves}
  \ForEach{leaf $\ell$}{
      $M_\ell \leftarrow
        dissimilarity(bel(\mathcal{H_\ell}^{0:t-1}),bel(\mathcal{F_\ell}^t))${\color{blue} \hspace{11.5em}(Eq. \ref{eq: similarity})}

      $C_\ell \leftarrow
        G(\mathcal{F}^t)$

      $J_\ell\leftarrow
        \lambda_1 C_\ell+
        \lambda_2 M_\ell$ {\color{blue}\hspace{21.5em}(Eq. \ref{eq: cost})}
  }

  \tcp{4) Execute robot move}
  $\ell^\star\leftarrow\arg\min_\ell J_\ell$;\;
  $a_r^{\,t}\leftarrow$ first robot step on path $\ell^\star$;\;
  execute $a_r^{\,t}$

  \tcp{5) Observe human and update belief}
  observe $a_h^{\,t}$\;
  $
    b_{t+1}(g)=
      \dfrac{\bar{\pi}_{g}(a_h^{\,t}\mid s_t,\mathcal H)\,b_t(g)}
            {\sum_{g'}\bar{\pi}_{g'}(a_h^{\,t}\mid s_t,\mathcal H)\,b_t(g')}; {\color{blue}\hspace{20em}
            \text{(Eq. \ref{eq:bayes_goal_update})}}
  $\
  $\mathcal H\leftarrow \mathcal H \cup \{a_h^{\,t}\}$\;
  $t\leftarrow t+1$
}
\end{algorithm}

\begin{figure*}
    \centering
        \includegraphics[width=\linewidth
    ]{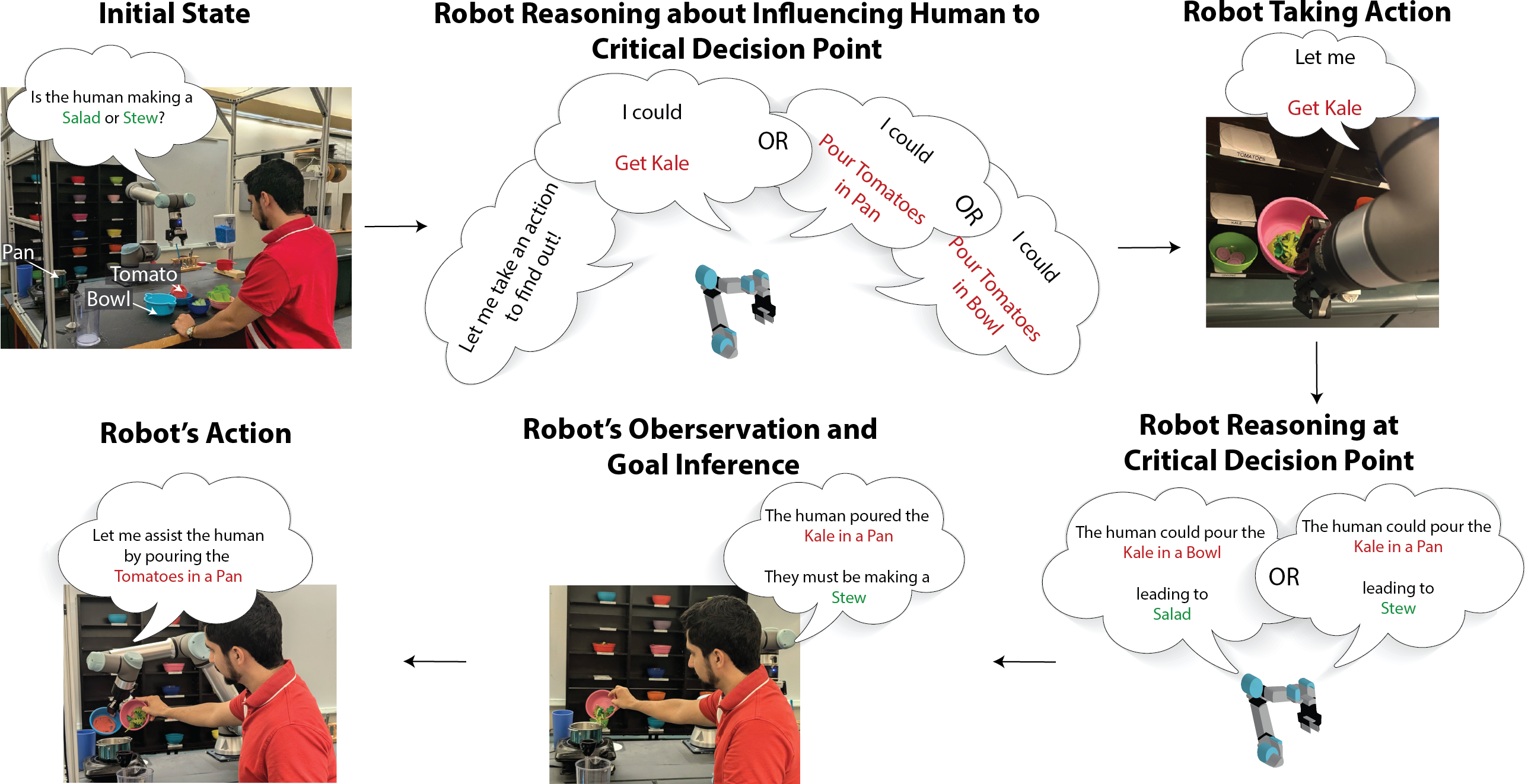}
    \caption{\textbf{Robot reducing uncertainty about the Human's Goal at a CDP:} As shown by the initial state, ingredients and utensils common to salad and stew are collaboratively brought by the human and the robot in the workspace. The robot reasons about possible next actions to influence the human to reveal their goal. The robot takes an action to complete gathering all the ingredients, rather than adding them to the pan or bowl, to observe the human in a CDP. This allows the human's next action (either adding the kale to the pan or the bowl) to reduce uncertainty about the goal. The human clarifies their goal through their action, and the robot takes action to support the human.}
    \label{fig:teaser}
\end{figure*}

\subsection{Approach}
\label{sec: collaborative_approach}

Alg. \ref{algo} describes our Receding Horizon Planning (RHP) approach for selecting the robot's action at a given timestep, such that the robot could influence the human to take more distinguishing actions at CDPs during human-robot collaboration. {\color{black}Fig. \ref{fig:teaser} visualizes a CDP in the collaborative cooking task where the prior actions are compatible with multiple possible meals, but the human’s next action, such as adding kale to the pan or bowl, distinguishes between candidate goals.}

First, we expand an RHP tree with the current interaction sequence (i.e., the history of actions leading up to the last action taken by the human) as its root. To do this recursively, we iterate through every leaf node in the tree, which is a sequence of possible future actions both agents can take from the root node of the tree. Then, for each valid action from the corresponding state, we execute that action and add the resultant state as a child node to the selected leaf node (lines 5-6). We continue this process until we reach depth $\mathcal D_H$. Then, we observe the history of actions taken only by the human, and use this to calculate the cost for every leaf node according to Eq. \ref{eq: cost} (lines 7-11). {\color{black}Finally, this online RHP procedure defines the robot’s action-selection rule where} we return the first action leading to the leaf node with the lowest cost (lines 12-13), observe the human's next action (line 15), and update the robot's belief (line 16). 

\begin{figure}[h!]
    \centering  
    \includegraphics[width=\linewidth]{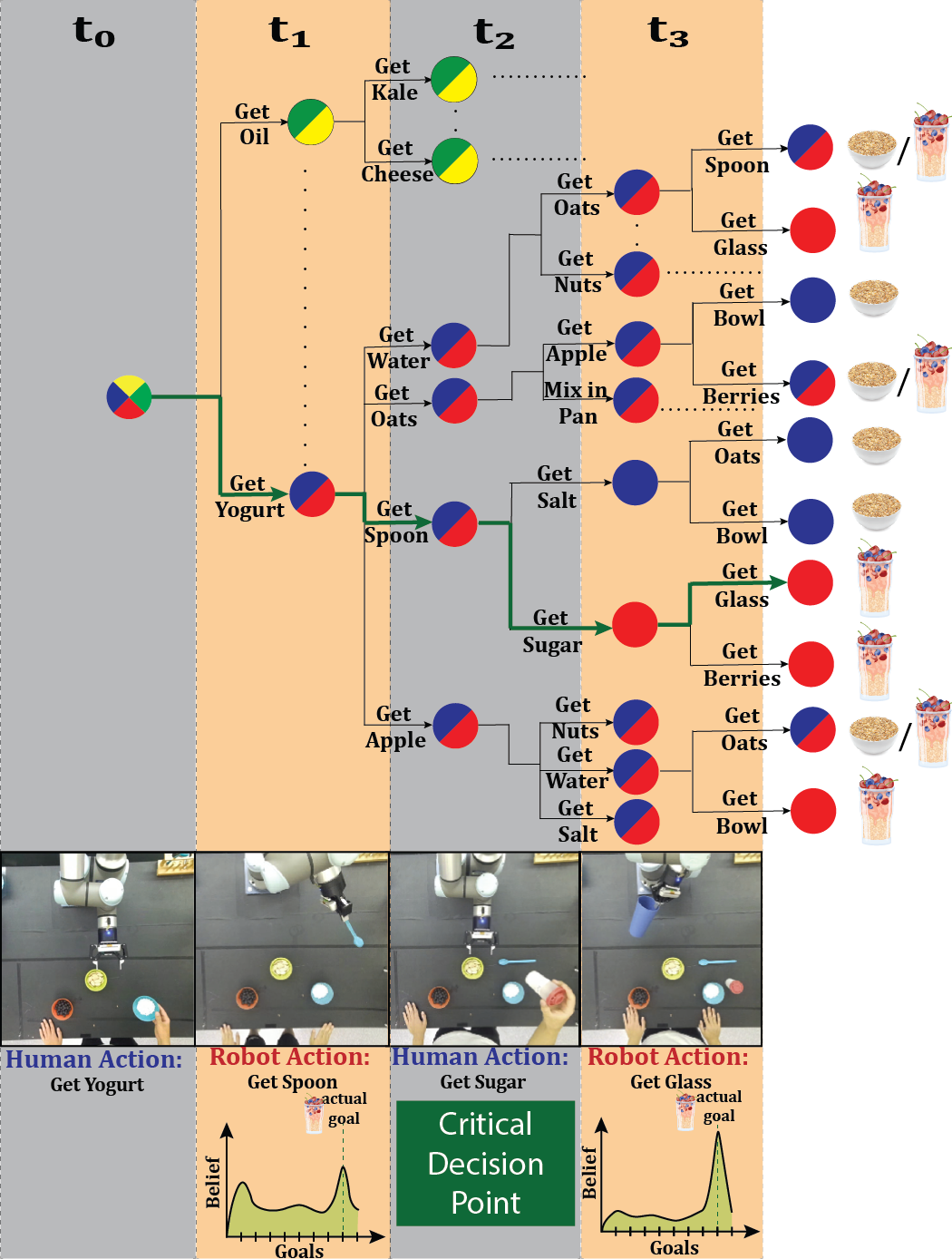}
    \caption{\textbf{Selecting actions to infer the human's goal at a CDP for a human-robot collaborative cooking task:} A Receding Horizon Planning (RHP) tree of all possible actions the human and the robot can take in a turn-based interaction is expanded, where the human takes the first turn. The circles correspond to states, and the arrows correspond to the possible action sequences taken at a given time step. The color of the circles indicates the possible goals that can be reached from a given state. The belief distribution is shown at steps $t_0$ and $t_2$, as the robot updates its belief over the human's goals when a human action is observed.  The green arrows show the actions that would optimize the cost function in Eq. \ref{eq: cost}. See the text for more details. 
}
    \label{fig:cdp_sequence}
\end{figure}

{\color{black} During collaboration, the robot’s objective is to take a supportive action. In our setting, this means selecting an action that} 1) is unlikely to be detrimental to achieving the human’s goal, even if the robot’s current goal estimate is wrong, and 2) is likely to update the robot’s belief about the human’s goal.


For the first consideration, we induce the robot to take general actions that are common for reaching multiple goals. {\color{black}More concretely, we consider an action to be general if it remains compatible with many possible goals in the policy bank.} Intuitively, as the task progresses, the robot’s tendency to act generically reduces the availability of general actions for the human, effectively pushing them to take specific actions that could clarify their true goal. {\color{black}For example, in the cooking domain, gathering a spoon or collecting water may support several recipes, while pouring a particular ingredient into a particular container commits more strongly to a smaller set of recipes.}
We formalize this via a generality score $G(\mathcal{F}^t)$, where $\mathcal{F}^t$ is a sequence of future actions taken from the state $s^t$. For a given sequence $\mathcal{F}^t$, the score is computed by estimating how often the policies in the bank $\mathcal{P}$ would result in the same sequence of actions as in $\mathcal{F}^t$.

For the second consideration regarding the belief about human goals, we calculate a dissimilarity measure that enables the robot to identify actions that maximize changes to this belief. We compute this dissimilarity measure as the disagreement of actions available from each state in the state space, as given in Eq. \ref{eq: cdp}. 
This dissimilarity measure takes into account the action history of an interaction up to state $s^t$, which we refer to as $\mathcal{H}^{0:t-1}$, and a possible sequence of future actions $\mathcal{F}^t$ from $s^t$:
\begin{align}
\label{eq: similarity}
    divergence :=dissimilarity(bel(\mathcal{H}^{0:t-1}),bel(\mathcal{F}^t))
\end{align}
where $bel(\cdot)$ corresponds to the likelihood over possible human goals induced by the policies in $\mathcal{P}$ given the input sequence, either past or future actions, 
{\color{black}and is computed using the Bayes filter from Eq.~\ref{eq:bayes_goal_update}.}
In this work, we calculate dissimilarity using cosine similarity; however, other measures, such as Wasserstein distance, could also be used as an alternative. Notably, this dissimilarity measure can be calculated online as the interaction progresses, if the state space is prohibitively large. 

Taking both considerations together, the robot ends up acting by choosing a branch of the RHP tree that minimizes the weighted combination of task and informativeness costs:
\begin{align}
\tiny
\label{eq: cost}
    C =  \underbrace{\lambda_1*G(\mathcal{F}^t)}_\text{task cost} + \underbrace{\lambda_2 * dissimilarity(bel(\mathcal{H}^{0:t-1}),bel(\mathcal{F}^t))}_\text{informativeness cost}
\end{align}
{\color{black}where, $\lambda_1$ and $\lambda_2$ are emperically determined hyperparameters.}
This cost is implemented in lines 16-17 of Alg. \ref{algo}, where the input sequence of past actions is computed in line 15, and the future possible actions are computed in lines 3 to 14.

Fig. \ref{fig:cdp_sequence} illustrates how our proposed approach works. A tree of all possible actions the human and the robot can take in a turn-based interaction is expanded, where the human takes the first turn. After $t_0$, the robot is already biased towards either the parfait or oatmeal being the human's goal, so it chooses a more general overlapping action (gathering a spoon) to influence the human into a CDP, which would be either gathering the salt or the sugar. At the CDP ($t_2$), the human would gather sugar, which disambiguates between the two most likely goals, leading the robot to update its belief and confirm the parfait as the goal. The robot can then align its actions with the actual objective, which in $t_3$ is gathering the glass. The green arrow shows the path that minimizes the cost. The robot would take the first action on this path, observe the environment, and reconstruct the tree at the next step.


\begin{table*}
\caption{List of Recipes that the Human and the Robot could collaboratively cook in our experimental setup \& the required ingredients.}
\label{tab: recipes}
\centering
\begin{tabular}{p{2cm}p{5cm}p{3.5cm}p{4cm}}
\hline
\textbf{Food Type} & \textbf{Basic Ingredients} & \textbf{Variants} & \textbf{Additional Ingredients} \\ \hline

Oatmeal     & water, oats, salt & 
Plain, Fruit, Chocolate, Fruit Chocolate & 
strawberries, blueberries, chocolate chips
\vspace{0.5em}
\\ 

Parfait     & yogurt, oats, sugar &
Plain, Fruit, Fruit Nut, Fruit Chocolate, Nut Chocolate, Fruit Nut Chocolate &
strawberries, blueberries, nuts, chocolate chips 
\vspace{0.5em}
\\ 

Pasta       & tomato, oil, pasta &
Tomato Cheese, Veggie, Veggie Cheese &
cheese, carrot, onion, broccoli 
\vspace{0.5em}
\\ 

Salad       & cucumber, lettuce, tomato, onion, kale, salt, oil &
Veggie, Pasta, Avocado, Nut Avocado, Fruit Avocado, Fruit Nut, Pasta Fruit Nut &
pasta, avocado, nuts, bananas, apple 
\vspace{0.5em}
\\ 

Smoothie    & water, yogurt, sugar &
Berry, Fruit, Chocolate, Chocolate Fruit, Kale Tomato, Kale Berry & 
strawberries, blueberries, bananas, apple, chocolate chips, kale, tomato 
\vspace{0.5em}
\\ 

Stew        & water, salt, onion, tomato, lettuce, cucumber, carrot, broccoli &
Veggie, Avocado, Cheese, Avocado Cheese &
avocado, cheese \\ 
\hline

\end{tabular}

\end{table*}

\subsection{Task and Problem Representation}

Collaborative cooking is a commonly used testbed for evaluating human-robot collaboration algorithms \cite{brawer2023interactive, carroll2019utility, van2022correct, goubard2023cooking, ghose2025ve, ghose2026open}. It is particularly interesting to demonstrate the efficacy of our approach because many overlapping sequences of actions can result in different meals being prepared. For example, chopping vegetables, boiling water, or mixing ingredients are common steps in various recipes, making it challenging to accurately infer the specific dish being made without additional distinguishing actions. {\color{black} In our setup, this ambiguity reflects the fact that the robot must collaborate from an incomplete task specification --- it is not given a complete step-by-step recipe in advance, and must infer the intended meal from the human’s actions during an interaction. This characteristic makes collaborative cooking a useful domain for evaluating whether a robot can effectively infer task intent and provide appropriate assistance under uncertainty.}

To evaluate our approach, we conducted experiments in both a simulation and a real-world setup where a human and a robot could collaboratively cook 30 possible recipes, taking turns performing one action each. 
We assume that the human and the robot can perform any required preparatory step, with each step taking the same time, regardless of who completes it. 
Our experimental setup consists of a robot and a user in a simplified kitchen environment (refer to Fig. \ref{fig:setup}) preparing a meal. The environment consists of a blender, a sink containing water, a pan on a stove, a serving bowl, a glass, a serving spoon, an eating spoon, a measuring cup, and a storage shelf containing ingredients for making the meals. The meals could be a type of pasta, stew, salad, oatmeal, smoothie or parfait. The details of recipes are given in Table \ref{tab: recipes}.

\begin{enumerate}[leftmargin=*]
    \item \textbf{Pasta:} The pasta sauce is prepared by blending tomatoes, pouring them into a pan with oil, and cooking it on the stove. The pasta is added to the stove and cooked. Variations include adding cheese or vegetables to the pasta. 
    \item \textbf{Stew:} The stew is prepared by filling a pan with water and adding salt, onion, tomato, and various vegetables to the pan. Variations include adding avocado, cheese, or both. The stew is cooked on the stove, mixed, and then served.
    \item \textbf{Salad:} The salad is prepared by adding ingredients like cucumber, lettuce, tomato, and onion into a bowl and mixing. Variations contain pasta, avocado, nuts, or fruits.
    \item \textbf{Oatmeal:} The oatmeal is prepared by filling a pan with water, adding oats and salt to the pan, and cooking. Variations include adding toppings like strawberries, blueberries, or chocolate chips. 
    \item \textbf{Smoothie:} The smoothie is prepared by adding water, yogurt, and sugar to the blender, followed by specific fruits or chocolate, depending on the recipe. The smoothie is blended, and the final product is poured into a glass.
    \item \textbf{Parfait:} The parfait is prepared by layering ingredients such as yogurt, oatmeal, and sugar in a glass. Variations may include adding fruits, nuts, or chocolate chips.

\end{enumerate}

The robot's state and action space are defined using the Planning Domain Definition Language (PDDL) \cite{aeronautiques1998pddl}. We used PDDLgym \cite{silver2020pddlgym} to dynamically update the environment's state based on actions taken by either the robot or the human. Both the human and the robot can take actions 

\begin{math}
    A = \{\texttt{gather(i,a)}, \texttt{pour(i,d,a)},  
    \texttt{mix(i,a)}, \\\texttt{cook(i,a)},
    \texttt{turn\_on(i,a)}, 
    \texttt{collect\_water(t,a)}, \\
       \texttt{blend(i,a)},
       \texttt{reduce\_heat(i,a)}, 
      \texttt{serve(i,a)}\}
\end{math}

\noindent
where $i$ is an item available in the interaction, $d$ is a container 
such as a bowl, glass, or pan, and $a \in \{HUMAN, ROBOT\}$ denotes the agent acting. The state space is described as a vector of PDDL literals where each element corresponds to a feature of the environment, including the presence of an item in the workspace or storage area, whether an item is in a container or appliance; the power state of the stove and blender; the state of an item (simmering, blended, cooked). We assume a deterministic environment and a fully observable state space.

\subsection{Experimental Setup}

The objective of both the physical and simulated collaborations is for a human and a robot to work together to prepare a meal, which could be a type of pasta, stew, salad, oatmeal, smoothie, or parfait (see Table \ref{tab: recipes}). While only humans know the specific meal being made, robots attempt to infer the intended meal by collaborating with humans and observing the sequence of actions they take in response to their own actions. We assume that the human always takes the first action in the interaction. Below, we detail our experimental setup for both the physical robot and simulation experiments.

\subsubsection{Simulation}
\label{sec: simulation setup}

\textbf{Modeling Human Behavior:} In the simulation, the human's policy bank is represented by ground-truth sequences of desired actions derived from corresponding Clique/Chain Hierarchical Task Networks (CC-HTNs) \cite{hayes2016autonomously}. These CC-HTNs model tasks that must be performed sequentially (e.g., the strawberry must be added to the blender before the blender is turned on) or can be done in an unordered fashion (e.g., the stew toppings can be retrieved in any order). 

\vspace{0.5em}
\noindent
\textbf{Approximating Belief Distribution: }Because our recipes are designed with many steps that can be performed in any order, the number of action sequences to complete a given recipe can become prohibitively large. This makes it challenging to calculate the belief distribution over possible goals after observing an action, as the lookup would need to account for all possible HTN combinations across all recipes. To address this, we approximate the belief distribution over goals using a transformer-based model 
 {\color{black}to approximate $bel(\mathcal H^{0:t-1})$ and $bel(\mathcal F^t)$ in Eq. \ref{eq: cost}}. 
 To train this goal prediction model, we sampled 10,000 valid action sequences for each of the 30 recipes (300,000 total action sequences across all recipes). The model employed a transformer architecture with multi-head attention \cite{vaswani2017attention}, comprising 4 layers and 8 attention heads, and was trained using cross-entropy loss to capture dependencies between actions within the sequence. The model produced a probability distribution over possible goals based on either a sequence of proposed actions ($bel(\mathcal{F}^t)$) or a sequence of actions performed by the human ($bel(\mathcal{H}^{0:t-1})$) in Eq. \ref{eq: cost}, with temperature scaling applied to control prediction confidence. The model was evaluated on a held-out validation set comprising 20\% of the dataset and produced a goal prediction accuracy of 83.86\%. This slightly low accuracy reflects the inherent difficulty of the task, as many potential goals share overlapping action sequences, making early goal prediction challenging.

\vspace{0.5em}
\noindent
\textbf{Expanding the RHP Tree:} Due to our large recipe set, as shown in Table \ref{tab: recipes}, we have a large number of possible actions valid at any given time step, making it computationally intractable to expand the RHP tree beyond a single timestep fully. To get around this limitation, we calculate the probabilities for each action, given a certain action history. We do this offline with an n-gram-based method, observing how often each action occurred with a certain prefix of n actions. Then, we only explore branches of the RHP tree ($\mathcal{B}$) corresponding to the top 5 most likely actions, which are returned by $get\_actions$ in Alg. \ref{algo} on line 7. This makes it computationally tractable for us to expand the tree to a depth of 4 ($\mathcal{D} = 4$).  {\color{black}The computational cost of RHP grows exponentially with the planning horizon. Since the planner expands a tree over alternating human and robot actions, the number of leaf nodes is $(O(B^{\mathcal{D}}))$. Each leaf is scored using the task and informativeness costs, so the total planning cost per timestep is $(O(B^{\mathcal{D}} C_{\mathrm{score}}))$, where $(C_{\mathrm{score}})$ is the cost of evaluating the cost function in Eq. \ref{eq: cost} at a leaf.}

 However, the n-gram-based approach also fails at a certain point, as the exponentially increasing number of unique action sequences for a given length means that we will eventually encounter a sequence not in our dataset. In this worst case, we fall back on our model's belief of the goal probability distribution and explore branches corresponding to actions that would help reach the most likely goals.

In our cost function described in Eq. \ref{eq: cost}, we selected $\lambda_1$ to be 1 and $\lambda_2$ to be 10. {\color{black}These weights control the tradeoff between selecting actions that remain useful across possible goals and selecting actions expected to reduce uncertainty about the human’s goal. We selected them empirically to prioritize informative interactions while still penalizing actions that would be unhelpful for task completion.}

\begin{figure}[t!]
    \centering
    \includegraphics[width=\linewidth]{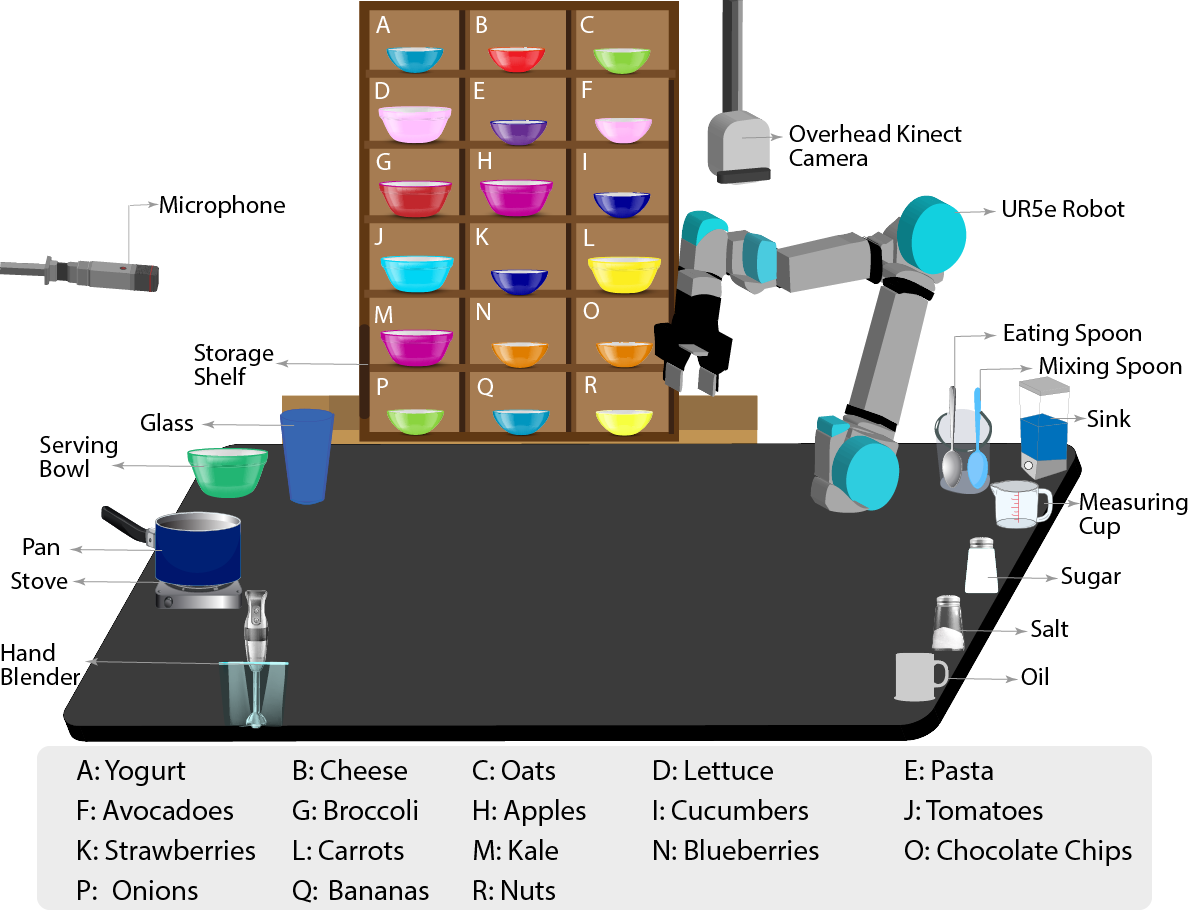}
    \caption{Setup for our Physical Robot Experiments for the Human-Robot Collaborative Cooking Task}
    \label{fig:setup}
\end{figure}

\subsubsection{Physical Robot}

As shown in Figure \ref{fig:setup}, our physical setup consisted of a UR-5e robot, an overhead Microsoft Azure Kinect camera \cite{smisek20133d} that monitored the position of ingredients in the workspace, and a microphone.
To avoid having to visually perceive the actions being performed by humans while they are doing them, humans were instructed to describe the actions they are doing in natural language. We then developed a simple language model that used fuzzy matching to map natural language to particular actions represented as PDDL literals. As shown in the supplementary video, realistic-looking artificial ingredients were used to ensure equipment safety, and liquids were replaced with colored beads. Similarly, all appliances were modified to turn on but not emit heat or have any sharp blades. The workspace in front of the robot contained a blender and a pan on a stove, both of which had been modified to enable the robot to manipulate them easily. To the robot's left, the water-dispensing sink, a stand containing the serving and the eating spoon, and a shelf containing the condiments, salt, sugar, and oil were placed. The serving bowl, the glass, and the measuring cup were located to the robot's right. The storage shelf containing the ingredients required to make the meals was behind the robot. Each ingredient was placed in a distinct colored container, which the robot could bring to the workspace whenever needed. Similar to Tilekbay et al. \cite{tilekbay2024expressedit}, we used the Contrastive Language Image Pretraining (CLIP) \cite{radford2021learning} model to localize objects in the robot's workspace based on predefined descriptive language prompts for each possible ingredient, to identify the region of interest containing the object and within that region of interest, we employed the Segment-Anything Model (SAM) \cite{kirillov2023segment} to predict a segmentation mask over the object of interest. Finally, we applied a heuristic grasp point prediction method using the segmentation mask, enabling the robot to manipulate the items in its workspace appropriately.

{\color{black}The physical robot setup was designed to instantiate the same symbolic task interface used in simulation. The perception and language-grounding modules mapped the real workspace into object locations, PDDL literals, and action labels used by the planner. This allowed us to evaluate whether our approach could be executed on a real robot in a controlled task setting, while keeping the task representation aligned with the simulation experiments.}

\begin{figure*}[t]
    \centering  
    \includegraphics[width=\linewidth]{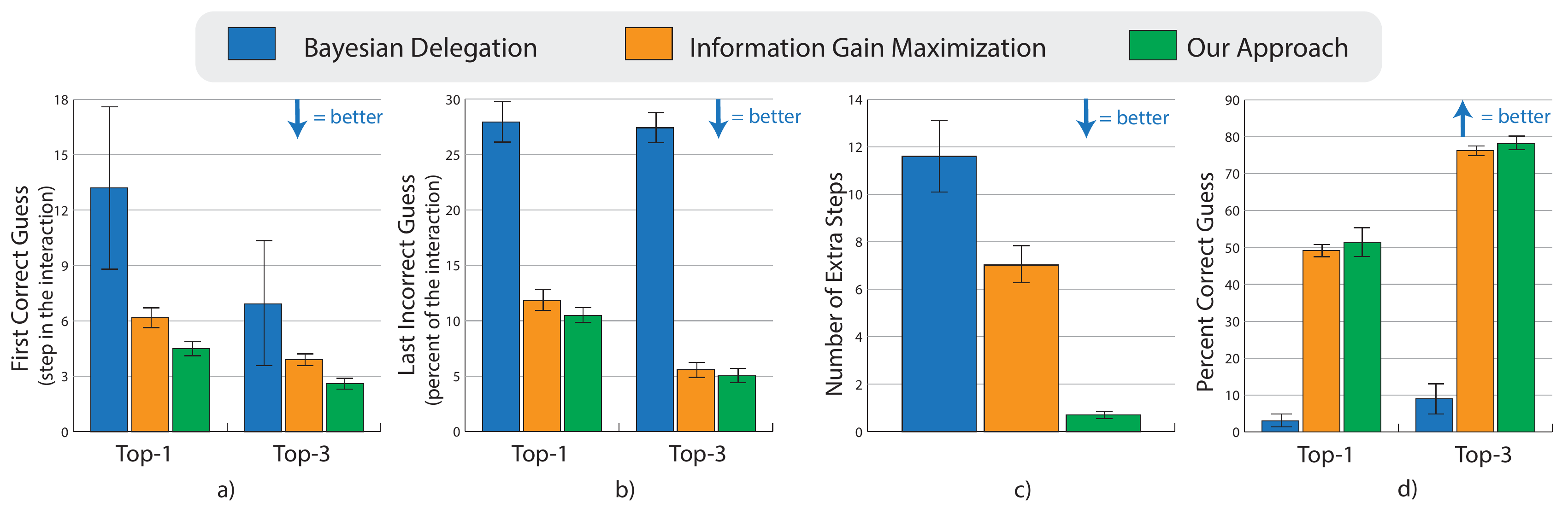}
    \caption{\textbf{Results of the Collaborative Cooking Evaluation: } Comparison of performance metrics between the proposed approach, vanilla CDP \cite{ghose2024planning}, Information Gain Maximization \cite{sadigh2016information}, and Bayesian Delegation \cite{wu2021too} in the collaborative cooking simulation. Metrics include
a) the first correct guess, b) the last incorrect guess, c) the number of extra steps, and d) the percentage of correct guesses. The means and standard deviations of the metrics are calculated over 30 recipes and evaluated across three trials.}
    \label{fig:results}
\end{figure*}

\subsection{Comparisons}

We compare the performance of our approach against the following two algorithms:

\subsubsection{Bayesian Delegation}

Bayesian Delegation \cite{wu2021too} uses Bayesian Inference to predict and coordinate actions in human-robot collaborative tasks. {\color{black}We use this approach as a belief-based baseline that updates its estimate of the human’s goal from observed actions while still selecting robot actions to move the collaboration forward. Unlike our approach, it does not explicitly select actions to guide the human toward CDPs.} Bayesian Delegation first determines the most likely task allocation from the action history, indicating which task each agent should work on. Then, it chooses each agent's action based on the goal it believes that the agent is trying to accomplish. 
Bayesian Delegation allows both agents to work on the same sub-task, which is not applicable in our setup. Moreover, it assumes that humans and robots act in parallel, which does not happen in our setup. We adapt the concurrent action execution in the original method to our turn-based setup by having the robot alternate actions with the human and expanding a tree of actions of depth 2 to simulate potential human responses. We consider the robot's actions and the human’s expected responses to calculate the Q values by giving a positive reward for each proposed action that aligns with the most likely recipe.

\subsubsection{Information Gain Maximization}
We use the information-gathering framework proposed in \cite{sadigh2016information} as another baseline in our setup. This approach enables a robot to actively probe a human's internal state by planning actions that maximize expected information gain. Following their method, we adapt it to our task by having the robot expand a receding horizon planning (RHP) tree of depth 4. Based on the original formulation of this approach, at each step, the robot optimizes a cost function that includes two terms: the first term maximizes the information gain of goal probabilities between the root node and the tree's leaf nodes, reducing ambiguity about the human's goal. The goal probabilities at the root and the leaf are calculated using the method described in Sec. \ref{sec: simulation setup}. The second term rewards the robot for taking actions aligned with the most likely goal and penalizes actions that deviate from it. The robot executes the first action that minimizes the combined cost in our implementation. It then allows the human to respond by taking the next valid action defined by the appropriate HTN, expands the tree again from the new state, and iteratively updates its goal probabilities.

{\color{black}In contrast to our CDP approach, this baseline intuitively selects actions expected to maximally shift the robot’s belief about human goals. CDP, by design, identifies informative states based on how much goal-conditioned policies disagree, regardless of the robot’s current belief, by computing informativeness directly from divergence in action choices. This baseline, on the other hand, defines informativeness as the expected change in the robot’s belief distribution. Unlike CDP, which assigns a fixed informativeness to each state, IGM’s measure varies with the robot’s current uncertainty, potentially causing the robot to overexplore states it is currently uncertain about, even if those states are unlikely to reveal the human’s true goal.}

\subsection{Metrics}

We compare the performance of our method against each baseline comparison for our proposed collaborative cooking task using four key metrics. 
\begin{enumerate}[leftmargin=*]
    \item \textbf{First Correct Guess: } The first timestep in an interaction when the robot correctly identifies the human's goal. 
    \item \textbf{Last Incorrect Guess: } The last timestep in an interaction when the robot wrongly identifies the human's goal. 
    \item \textbf{Number of Extra Steps: } Number of extra steps taken by both agents to complete the interaction compared to the number of ground-truth steps from the corresponding HTN. 
    \item \textbf{Percentage of Correct Guesses: } Percentage of timesteps during an interaction when the robot correctly identifies the human's goals. 
\end{enumerate}
We report each method's top-1 and top-3 accuracy for the first correct guess, the last incorrect guess, and the percentage of correct guesses.

\begin{figure*}[ht]
    \centering  
    \includegraphics[width=\linewidth]{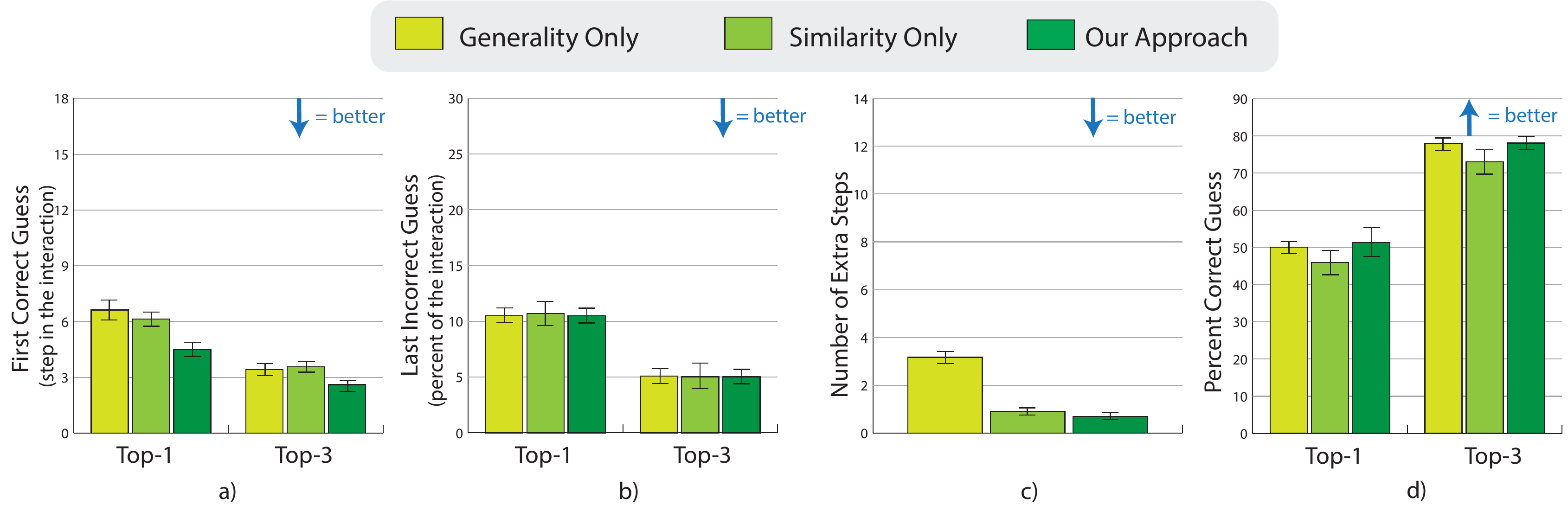}
    \caption{Ablation Study over the components of the Cost Function for the Collaborative Cooking Task}
    \label{fig:ablation_results}
\end{figure*}

\begin{figure*}[ht]
    \centering  
    \includegraphics[width=\linewidth]{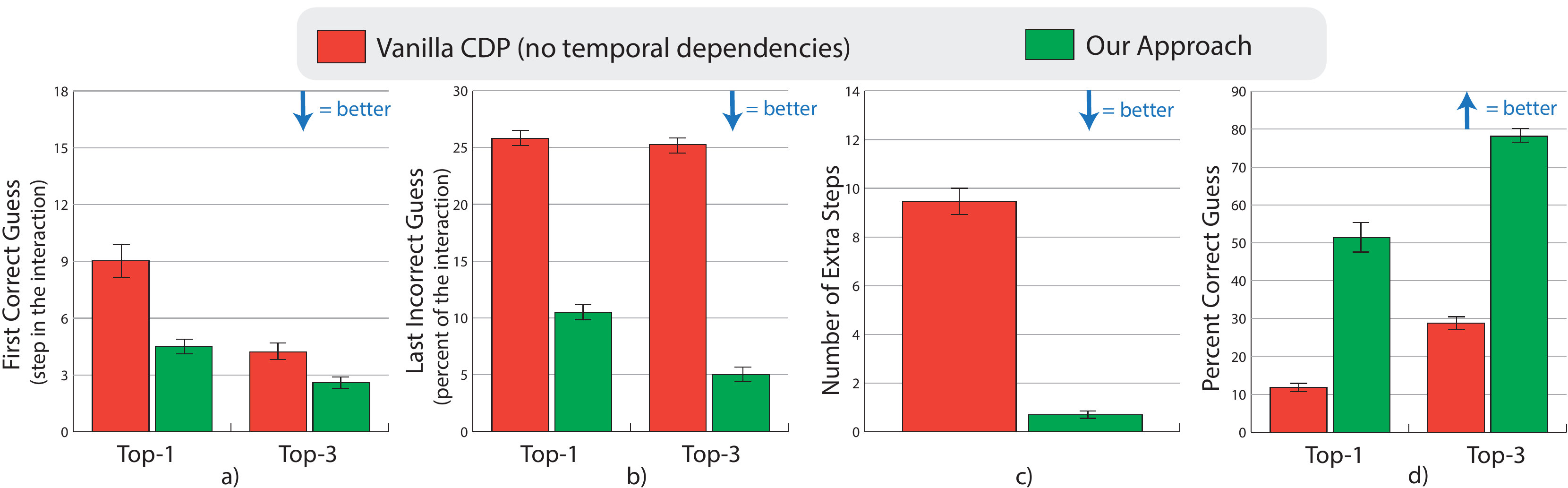}
    \caption{Ablation Study over using temporal modeling for computing CDPs for the Collaborative Cooking Task}
    \label{fig:time_ablation_results}
\end{figure*}

\subsection{Simulation Results}


Fig. \ref{fig:results} compares our proposed approach with the Bayesian Delegation \cite{wu2021too} and Information Maximization \cite{sadigh2016information} approaches across several key metrics in a simulated collaborative cooking task. The means and standard deviations of the metrics are calculated over 30 recipes and evaluated across three trials. When following the ground truth, each interaction sequence, on average, involves 18.7 steps, setting a baseline to assess the robot's performance.

Fig. \ref{fig:results} a) shows that our approach consistently identifies the human’s goal earlier in the interaction than all baseline approaches. The first correct guess typically occurs in the first third of the sequence, well before the interaction’s midpoint. However, this metric can be somewhat noisy, as all methods might occasionally predict the correct goal by chance early in the interaction without resolving uncertainty about the goal.

\begin{figure*}[]
    \centering
        \includegraphics[width=\linewidth]{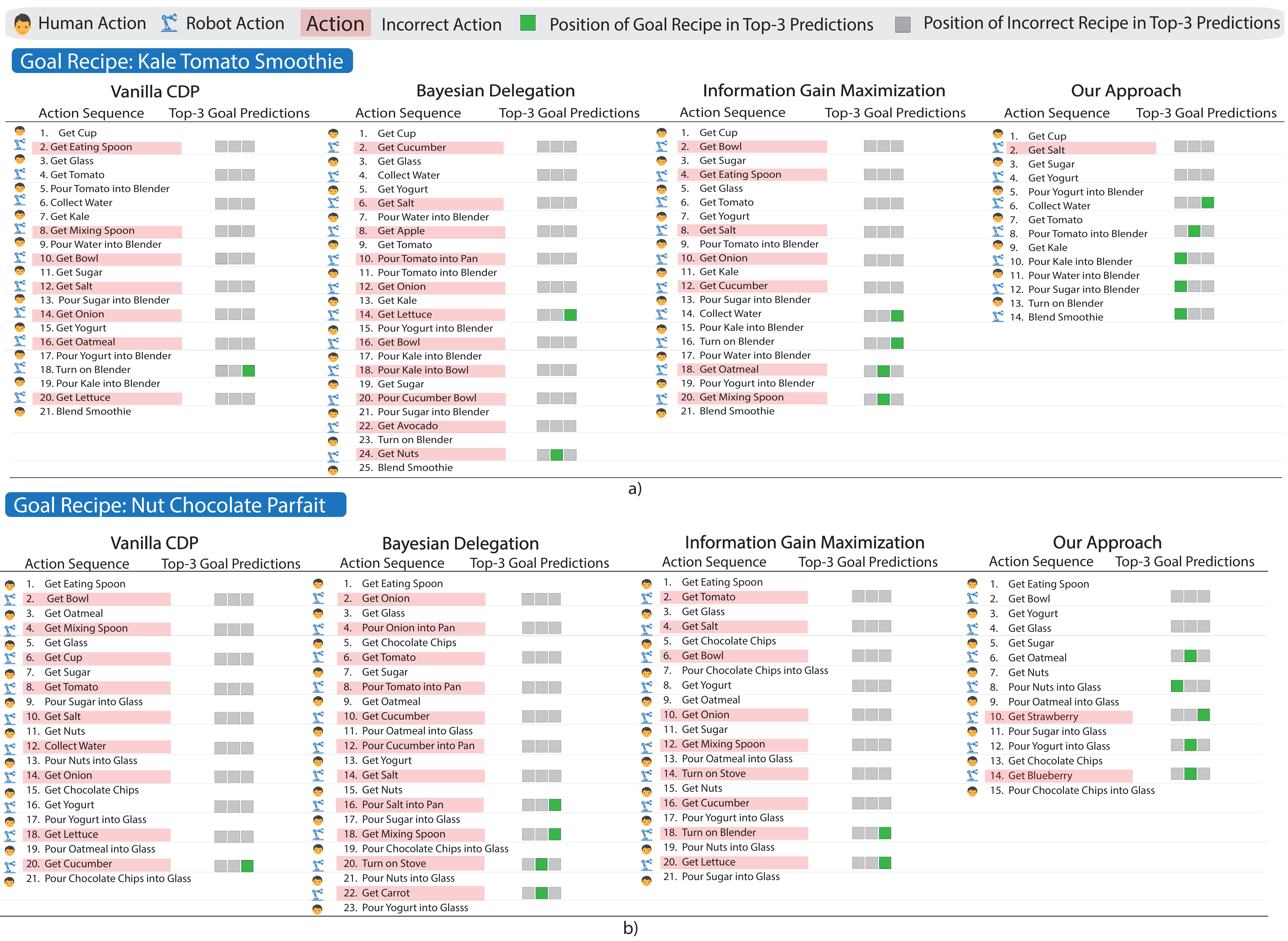}
    \caption{\textbf{Collaborative Cooking Case Study: }Results of the Case Study with the Physical Robot running Bayesian Delegation, Vanilla CDP (history agnostic), Information Gain Maximization, and Our Proposed Algorithm: For Case Studies, a) and b), a group of columns represents the method. For each group, the first column depicts the agent taking action, the second column depicts the action sequences taken by the respective method, and the third column denotes the presence and position of the goal recipe in the top-3 most probable recipes from the goal prediction model (left-most being the most probable recipe).}
    \label{fig:case_study}
\end{figure*}

Therefore, we compute the last incorrect guess metric, as it marks the point in the interaction where the robot no longer makes erroneous predictions about the human’s goal. Fig. \ref{fig:results} b) shows that our approach generally achieves this around the midpoint of the sequence, earlier than all baselines. This earlier resolution of uncertainty indicates that by the time half the sequence is completed, our method has effectively understood the human's intentions, leading to a higher percentage of correct guesses as shown by Fig. \ref{fig:results} d).

Fig. \ref{fig:results} c) shows the total number of extra steps both agents took to complete the collaboration. If the robot performs every action incorrectly, the interaction could take up to twice as long, as the human would need to perform every correct step towards task completion. Conversely, if the robot correctly assists the human, the task can be completed in the baseline of 18.7 steps, the average length of a ground truth sequence. The number of extra steps in our approach is minimal, averaging 0.70 steps, compared to 7.02 steps for the information gain maximization approach, 9.46 steps for the vanilla CDP approach, and 11.0 steps for the Bayesian Delegation approach, showing that our approach can lead to efficient human-robot collaboration.

\noindent
\subsection{Ablation Study}

\subsubsection{Cost Function Ablation}
{\color{black}Fig. \ref{fig:ablation_results} shows the results of the ablation study used to evaluate the contribution of the generality and similarity terms in Eq. \ref{eq: cost}. The generality-only condition prioritizes actions that remain compatible with many possible goals, while the similarity-only condition prioritizes actions that are expected to change the robot’s belief about the human’s goal. Our full method performs better than either individual term across all metrics, suggesting that both components are necessary. This supports our use of the weighted cost function, where $\lambda_1$ and $\lambda_2$ control the tradeoff between task progress and information gain.}

 When generality is considered individually, we observed that the robot tended to take very general actions, often avoiding task-specific actions. This behavior caused the robot to deviate from the task at hand, leading to unnecessary steps to complete the task, as illustrated in Figure \ref{fig:ablation_results}c). Notably, since the robot is inclined to undertake actions that correspond with a broader range of goals by prioritizing generality, it performed well in accurately predicting the correct goal, as shown in Figure \ref{fig:ablation_results}d).
 
On the other hand, when the similarity term is considered in isolation, we found that the robot effectively selects the most beneficial actions for the human in the moment. However, it struggles to guide the human toward CDPs, which are essential for quickly clarifying their goals. This limitation is evident in Figure \ref{fig:ablation_results}d), where the robot performs poorly in goal prediction. Conversely, it takes fewer unnecessary steps, as depicted in Figure \ref{fig:ablation_results}c).

\subsubsection{Time Dependency Ablation}

In this ablation (Fig. \ref{fig:time_ablation_results}), we isolate the effect of incorporating interaction history into the 
{\color{black}informativeness score calculation (Eq: \ref{eq: similarity}) to establish a ``history‐agnostic" CDP baseline, in which each state’s informativeness is computed in isolation from any prior actions (this baseline is called ``Vanilla CDP" in the rest of the Section).}  At every time step, this approach evaluates the divergence among goal‑conditioned policies at the current state, chooses the move that leads to the highest immediate information gain, observes the human’s response, and updates its belief, never allowing past observations to influence which states are deemed most revealing.  By contrast, our approach conditions the informativeness calculation on the entire interaction history $\mathcal H$. As Fig.\ref{fig:time_ablation_results} shows, this temporal reasoning enables the robot to correctly identify the human’s goal significantly earlier in the interaction and to complete the task in far fewer extra steps than the history‑agnostic CDP baseline.

\subsection{Real Robot Results}

We conducted two proof-of-concept case studies using a real robot and a human collaborator (with one of the researchers acting as the collaborator using the setup shown in Figs. \ref{fig:teaser} and \ref{fig:setup}) to compare the performance of Bayesian Delegation \cite{wu2021too}, information-gain maximization \cite{sadigh2016information} and vanilla CDP \cite{ghose2024planning} methods with our proposed approach in collaborative cooking tasks as shown in Fig. \ref{fig:case_study} (refer to the Supplementary Video). For each step the robot took in all four methods, we evaluated the predictions of our goal predictor model to determine whether the correct goal recipe is among the top three most probable options. To ensure a fair comparison of the methods, humans always took the same first action, preventing any revealing actions at the outset. 

We found that our method strategically selected general actions common to most recipes to influence the human collaborator to perform distinguishing actions early, reducing uncertainty about the human's goal. This allowed the robot to become certain sooner and avoid incorrect actions. In contrast, Bayesian Delegation relied on noisy initial goal estimates, leading to early mistakes that compounded and made it difficult for the goal predictor to identify the correct recipe by the end. Vanilla CDP identifies Critical Decision Points (CDPs) based solely on the immediate information gain at a given state, without considering the action history or the long-term impact of prior actions, which is shown to be problematic given the sequential nature of the task. Finally, we find that Information Gain Maximization fails to balance exploration (gathering information to clarify the goal) with exploitation (taking actions that align with the most likely goal). 
Consequently, for each of our three baseline approaches, the human performed most steps in the task.                                                

Fig. \ref{fig:case_study}a) shows our first case study where our goal recipe is the Kale Tomato Smoothie; our approach influenced humans to add distinctive ingredients early by performing general actions like getting the yogurt and collecting water. The robot became certain of the goal by step 10 by just taking one extra step at the beginning of the interaction. Bayesian Delegation, relying on uncertain estimates, led the robot to make mistakes such as pouring tomato into the pan instead of the blender, adding unnecessary ingredients like lettuce, and taking ten extra steps. 
Vanilla CDP fails to consider the entire action history, leading the robot to use combinations of actions that do not contribute to any recipe in the Goal Bank, e.g., Get Glass and Get Mixing Spoon, which never occur together in any recipe. This led the robot to take seven extra steps. 
Information Gain Maximization tries to narrow down the goal distribution without considering whether the goal it is converging to is correct or not, which leads it to take overly specific actions that are not helpful to the recipe, such as Get Onion, leading the robot to take seven extra steps.

Fig. \ref{fig:case_study}b) presents our second case study, where the goal recipe is Nut Chocolate Parfait. Our approach began with general actions like retrieving the bowl and glass and influencing the human to add yogurt, nuts, and chocolate chips early, key steps in making the parfait. As a result, the robot quickly became certain of the goal, with the correct recipe consistently appearing in the top 3 predictions from step 6 onward. Although our method took a few extra steps due to confusion between the Fruit Nut Chocolate Parfait and the Nut Chocolate Parfait because the robot erroneously fetched strawberries and blueberries, it only required three extra steps. In contrast, Bayesian Delegation caused the robot to make irrelevant actions like retrieving tomato, cucumber, and onion, leading to ten extra steps and fewer correct guesses.
Like in Fig. \ref{fig:case_study}a), Vanilla CDP takes combinations of actions that can never plausibly occur together, such as Get Glass and Get Salt. This leads the robot to take eight additional steps.
Information Gain tries to acquire as certain an estimate of the human's goal as soon as possible, which leads to it taking incorrect actions with relatively low generality very early (such as Get Tomato), which leads the robot to take eight extra steps.

\subsection{Comparative Analysis of Performance of CDPs in the Collaborative Task}

\subsubsection{Bayesian Delegation}
Bayesian Delegation \cite{wu2021too} struggles when dealing with tasks involving overlapping action sequences that lead to different goals. When different goals share many of the same actions, this approach struggles to differentiate between them because the observed actions do not provide enough discriminatory power, causing the belief to remain spread across multiple possible goals. This results in delayed disambiguation, as the Bayesian approach accumulates evidence slowly, making it harder for the robot to infer the correct goal promptly and confidently. Moreover, the probability mass tends to be diluted across several goals rather than concentrating on the correct one, making it difficult for the system to favor the correct goal even after multiple observations. As observed in our experiments, these limitations lead to slower and less accurate goal inference.

\subsubsection{Information Gain Maximization}
Information Gain Maximization \cite{sadigh2016information} focuses on reducing uncertainty at each step by selecting actions that maximize immediate information gain. However, this approach is shortsighted for tasks where multiple goals share the same action sequences because the robot prioritizes actions that provide immediate information. Intuitively, this method fails to balance exploration (gathering information to clarify the goal) with exploitation (taking actions that align with the most likely goal). By always taking the action that provides the most information immediately, it may be pursuing the wrong goal, when it should be taking more general actions that provide less information but help support the human's goal.

\subsubsection{Vanilla CDP}

To isolate the effect of incorporating interaction history into our approach, we compare our method against vanilla CDP \cite{ghose2024planning}.
Vanilla CDP considers only the current state without past actions. This leads the robot to take implausible actions, as it ignores the context in which human actions were performed. The issue is especially problematic when multiple action sequences overlap, making context crucial for distinguishing between goals. Without an action history, the robot’s mistakes compound, making it harder to infer the human’s goal. 
Also, since whether a state is informative about a human's goal depends on when it occurs, ignoring temporal context weakens the robot’s ability to identify key moments that reveal human intent.

\subsubsection{Our Approach}
In contrast to the above baseline methods, our approach influences human actions actively to reveal their goal and addresses the challenges faced by other methods.

We address the issue of overlapping action sequences faced by Bayesian Delegation by guiding the human towards more actions that reduce the robot's uncertainty about the human's true goal. This strategy reduces the impact of overlapping sequences by incorporating generality in the cost function, which steers the human towards taking distinguishing actions while also increasing the likelihood that, even if the robot's belief about the human's goal is incorrect, the chosen action remains on policy, ensuring that the robot's actions are robust to uncertainties in goal prediction.

Our approach addresses the issues faced by Information Gain Maximization by incorporating similarity into our cost function. By doing this, we encourage the robot to take actions that are likely to be relevant to the human's goal, rather than the action that maximally narrows down the belief distribution over the human's goals, regardless of whether the goal it is converging to is correct.

Finally, unlike Vanilla CDP, our approach considers the entire action history when selecting new actions, which prevents us from taking implausible combinations of actions, which are actions that are unlikely to be taken together in the context of the policy bank $\mathcal{P}$. Using our approach, the robot aims to steer the collaboration toward taking actions that are unlikely to be detrimental to achieving the human's goal or significantly alter the robot's belief over human's goals. In effect, it enables humans to take actions that reduce the robot's uncertainty about its true goal earlier in the interaction than vanilla CDP. 

\section{Scenario 2: Influencing the Human to Reveal their Goal in a Competitive Human Robot Task}

\subsection{Approach}

\begin{algorithm}[t]
\small
\caption{Scenario 2: Selecting Action at Timestep t to Influence Humans to CDPs by expanding an RHP tree for Human-Robot Competition}
\label{alg:competition}
\KwIn{%
  Goal set $\mathcal G=\{g_1,\dots,g_n\}$ with policy banks
  $\mathcal P_g=\{\pi_{g}^{1},\dots,\pi_{g}^{n_g}\}$;          
  Pre‑computed CDP set $S_{\mathrm{CDP}}(\varnothing)$;               
  planning horizon $D_H$; weights $(\lambda_1,\lambda_2)$;         
  robot action space $\mathcal A_r$}
\KwOut{Next seeker action $a_r^{\,t}$}

$b_0(g)\gets 1/|\mathcal G|$,\;
$t\gets0$

\While{hider not caught}{
  \tcp{1) Sense current state}
  observe $s_t=(s_w,s_h,s_r)$

  \tcp{2) Build depth‑$D_H$ tree}
  Alternate levels of
  \begin{itemize}[leftmargin=*,itemsep=0pt]
    \item \emph{Seeker branch}: enumerate $a_r\in\mathcal A_r$
    \item \emph{Hider branch}: sample
          $a_h\sim\bar{\pi}_{\hat g}(a_h\mid s_t)$ \\where
          $\displaystyle
            \bar{\pi}_{g}(a\mid s)=\frac1{|\mathcal P_g|}
            \sum_k\pi_{g}^{k}(a\mid s)$ \hspace{0em} {\color{blue}(Eq. \ref{eq:marginal_pi})}
  \end{itemize}
  Store leaves $\ell$ with state $s_\ell$ and future sequence
  $\mathcal F_\ell$

  \tcp{3) Score each leaf}
  \ForEach{leaf $\ell$ generated under goal $g$}{
      $C_{\text{CDP}}(\ell)=
        \min_{s_{\text{crit}}\in S_{\mathrm{CDP}}}
          \mathrm{dist}(s_{h,\ell},s_{\text{crit}})$  \color{blue} \hspace{10em}(Eq. \ref{eq:cdp_comp})}

      $C_{\!\mathrm{task}}(\ell)
          = C_{\!\mathrm{task}}(s_\ell)$

      $J_\ell=
        \lambda_1\,C_{\!\mathrm{task}}(\ell)+
               \lambda_2\,C_{\text{CDP}}(\ell)
              \dfrac{1}{b_t(g)}$ { \hspace{20em}\color{blue}(Eq. \ref{eq:rhp_cost})}
  }

  \tcp{4) Execute best seeker move}
  $\ell^\star=\arg\min_\ell J_\ell$;\;
  $a_r^{\,t} \leftarrow$ first seeker action on path $\ell^\star$;\;
  execute $a_r^{\,t}$

  \tcp{5) Observe hider and update belief}
  observe $a_h^{\,t}$\;
  $
    b_{t+1}(g)=
      \frac{\bar{\pi}_{g}(a_h^{\,t}\mid s_t)\,b_t(g)}
           {\sum_{g'}\bar{\pi}_{g'}(a_h^{\,t}\mid s_t)\,b_t(g')}; {\color{blue}
            \text{(Eq. \ref{eq:bayes_goal_update})}}
  $\
  $t\gets t+1$
\end{algorithm}

{\color{black}In this scenario, we consider a different instantiation of the problem: a competitive environment in which the robot seeks to infer the human’s goal under partial observability. Unlike the collaborative setting where the robot can condition its strategy on the complete interaction history $\mathcal{H}$, here the history may be unavailable or unreliable due to limited sensing or occlusion. This changes the nature of the inference problem -- requiring the robot to reason from immediate state information alone and adapt its strategy accordingly. However, this enables the robot to pre-compute CDPs if the state space is not prohibitively large. }

\begin{figure}[t!]
    \centering
    \includegraphics[width=\linewidth]{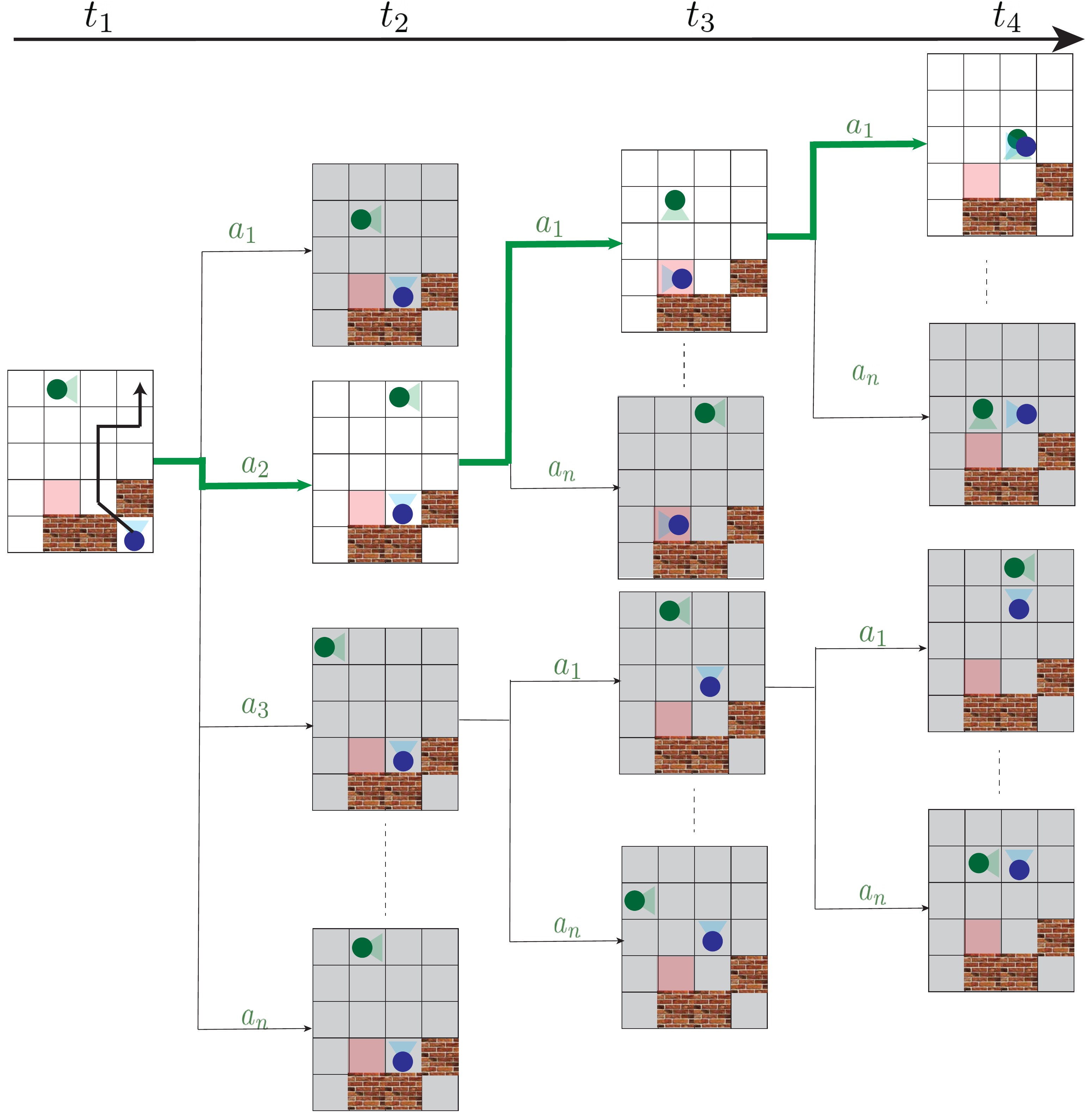}
    \caption{\textbf{Example RHP tree rollout for influencing the hider in a competitive hide-and-seek game:} {\color{black}The green autonomous seeker robot and blue human-operated hider robot play hide-and-seek on a grid with brick-wall obstacles. The seeker uses RHP to steer the hider toward a Critical Decision Point (the red cell), where the hider’s next move can reveal its hidden strategy and speed capture. The green action labels $(a_1, a_2, a_3, \ldots, a_n)$ denote candidate seeker actions in the RHP tree. The black arrows inside each grid show the hider’s movement under a possible policy from the policy bank. At each timestep, the seeker rolls out possible future interactions to horizon $(t_4)$, considering both its own candidate actions and the hider’s possible policy-conditioned responses. At $(t_1)$, both agents follow their underlying plans. When the hider spots the seeker at $(t_2)$, it detours evasively at $(t_3)$, then resumes its original route at $(t_4)$, landing on the CDP unobserved and allowing the seeker to infer its strategy and catch it. The rollout identifies the optimal seeker action sequence $(a_2 \to a_1 \to a_1)$ shown by the green arrows; by RHP, the seeker executes only the first move $(a_2)$ before replanning. See the text for more details.}}
    \label{fig:mpc}
\end{figure}


Alg. \ref{alg:competition} summarizes our approach for a robot to influence a human towards Critical Decision Points in a competitive task under partial observability. Similar to the human-robot collaboration scenario described in Sec. \ref{sec: collaborative_approach}, at a given time $t$, the robot rolls out all possible future actions that can be taken by the human per the policy bank $\mathcal{P}$ along with its own responses to the human's possible actions. These steps are repeated to grow the search tree to a certain horizon $\mathcal D_H$. For every leaf node in the tree that results from the human acting according to a policy $\pi_i$, the robot computes a cost $\mathcal{C})$ that describes how good that specific future is based on 1) the
distance between the projected human position at time $t + \mathcal{D}$ and the closest precomputed CDP ($s_{crit}$) as a measure of informativeness similar to Eq. \ref{eq: similarity},
\begin{equation}
    \label{eq:cdp_comp}
    divergence := C_{CDP} = dist(s_h^{t+\mathcal D_H}, s_{crit})
\end{equation}

\noindent
and 2) other task-related costs that support the robot's competitive objectives, like the distance between the human and the robot:
\begin{align}
    \mathcal{C} =   (\underbrace{\lambda_{1}C_{task}}_\text{task cost} + \underbrace{\lambda_{2}C_{CDP}}_\text{informativeness cost})\dfrac{1}{b_t(g)}
    \label{eq:rhp_cost}
\end{align}
where $dist(s_h^{t+\mathcal{D}}, s_{crit})$ is the shortest path between the human and the closest Critical Decision Point $\mathcal{D}$ steps in the future when the human acts according to $\pi_i$ and the robot takes corresponding actions. 
{\color{black}The informativeness cost $C_{CDP}$ serves the same function as CDP computation in the collaborative scenario, but is pre-computed rather than being dynamically inferred.} $\lambda_{1}, \lambda_{2}$ are weights to balance the task and informativeness costs. The cost $\mathcal{C}$ is scaled by the likelihood of the human policy $b_t(g)$, where 
$b_t(g)$ is the robot’s belief that the human’s hidden goal is $g$ (updated via the Bayes filter in Sec.\ref{sec: belief} with the
same marginal likelihood $\bar{\pi_g}$). 
such that policies that are less likely will result in higher costs. 
Finally, the robot selects its best response action $a_r^*$ at time $t$ as the action that minimizes the cost function $\mathcal{C}$ for all possible $\pi_i \in \mathcal{P}$ over the horizon $\mathcal D_H$. This allows it to influence the human toward Critical Decision Points, because the cost depends on where the human ends up after $\mathcal D_H$ steps in the future.

Fig. \ref{fig:mpc} illustrates how our proposed approach works. The green and the blue agents are playing a hide-and-seek game, where the green agent (the seeker) is trying to influence the blue agent (the hider) to a Critical Decision Point (the red cell) using RHP to infer its policy and therefore catch it faster. The brick walls represent obstacles. The green agent imagines all possible future actions, assuming the blue agent follows a fixed policy, and selects the most optimal action in the next timestep to lead to the most favorable outcome at horizon $\mathcal D_H=t_4$. The black arrow at $t_1$ represents the underlying strategy. Both agents' fields of view are restricted to all the cells in the direction they face. At $t_2$, when the blue agent detects the green agent, it abandons its underlying strategy at  $t_3$ to perform an evasive maneuver, then resumes its original path once out of the green’s field of view at $t_4$. This allows the green agent to observe the blue agent's underlying policy at a critical decision point at $t_3$ without being observed, and ultimately catch it in $t_4$. Note that the green arrow denotes the optimal next action for the green agent. After rolling out all possible futures, the green agent determines its optimal sequence of actions to be $a_2 \rightarrow a_1 \rightarrow a_1$ (depicted by the green bold arrows). Since the green agent performs RHP, it takes the first action $a_2$ from the sequence and constructs the tree from the next state.


\begin{figure*}
    \centering
    \includegraphics[width=\linewidth]{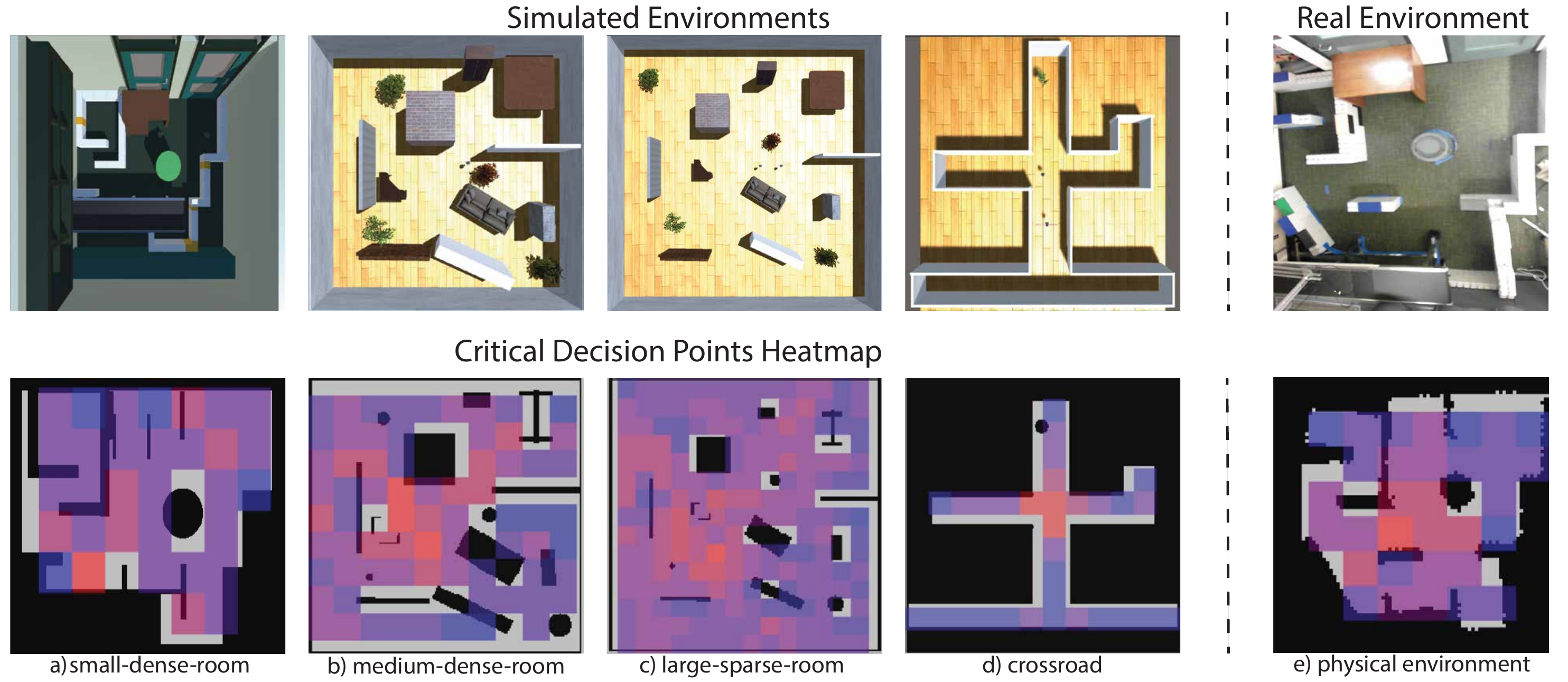} 
    \caption{\textbf{Test Environments and Computed CDPs for the Competitive Hide and Seek Task: }(Top) Photo-realistic Simulated and real-world environments for the hide-and-seek task. (Bottom) Heatmaps of Critical Decision Points for corresponding environments. States closer to red denote the most Critical Decision Points, and states closer to blue denote the least critical points. Note: Although the figure displays Critical Decision Points for every state, typically, it is sufficient to compute them in a small region around the robot.}
    \label{fig:environment_CDP}
\end{figure*}

\begin{figure}
    \centering
    \includegraphics[width=\linewidth]{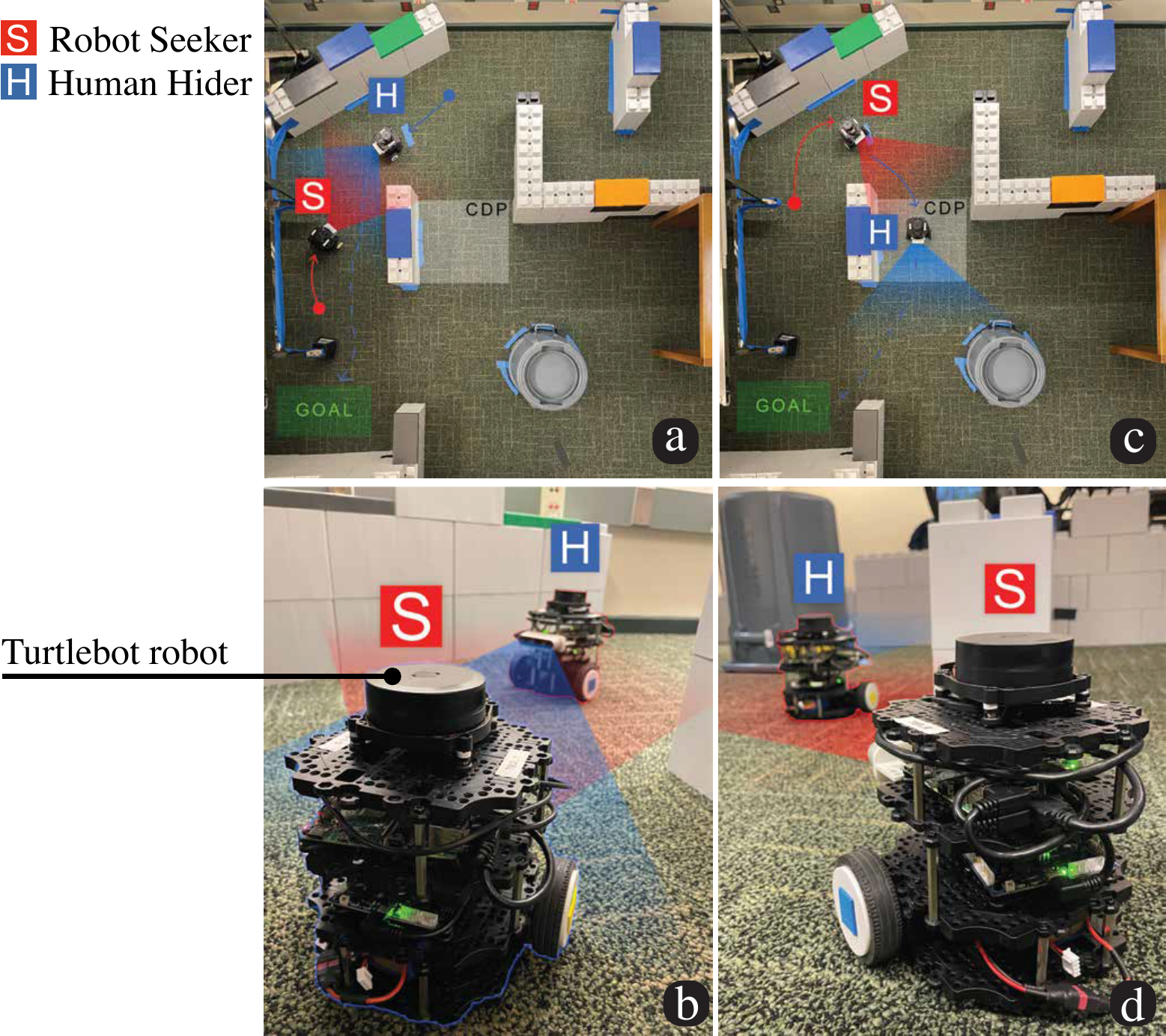}
    \caption{\textbf{The robot seeker (S) influences the human hider (H) towards a Critical Decision Point (CDP) in a hide-and-seek game:} a), b) The human-controlled hider sees the robot seeker on its way to its goal while following its high-level strategy. c) The robot seeker actively influences the human-controlled hider to escape towards a CDP (gray area). d) When the hider is at the CDP, the robot seeker optimizes for not being seen by the hider. So, the hider continues to navigate toward its goal. The robot seeker sees the hider take action toward its goal and identifies its strategy.}
    \label{fig:realrobot}
\end{figure}

\subsection{Task and Experimental Setup}

\subsubsection{Task}
We use the game of hide and seek as our experimental setup, as shown in Fig. \ref{fig:realrobot}. The game allows the construction of arbitrarily complex environments, requires the agents to take a large variety of environment-dependent strategies, and requires that agents not communicate about their strategies \cite{baker2019emergent}. {\color{black}In this setting, the hider’s policy bank defines a set of possible routes or strategies that the hider may privately follow during an episode. The seeker knows the space of possible strategies but does not know which one the hider is executing, and must infer it from partial observations.} {\color{black}Moreover, we believe it is a good testbed to evaluate our approach in a partially observable, discretizable environment, in a reasonably small state space, to showcase how CDPs can be pre-computed, and how humans can be influenced by CDPs to reveal their strategies. }

We developed the hide-and-seek task in a photo-realistic simulation and the real world. In both settings, we embodied the agents with Turtlebot3 robots, so there is no advantage due to the different agent morphology in the game. In other words, when humans played the game, they did so by controlling the TurtleBot3 robot using a joystick, 
and its perception and control abilities thus limited them. The robots have a forward-facing camera, providing a first-person view of the environment. The placement of objects in the environment and agents' access to visual information through their cameras make the hide-and-seek game partially observable for both agents. 

The photo-realistic simulation environments were created by modifying the SEAN 2.0  simulator \cite{tsoi2022sean} to support multiple robots.  In simulation, an autonomous agent assumes the role of the seeker, while a simulated human controls the hider. In all experiments, both agents access similar sensor data and have the same form factor. 
For real-world experiments, we set up a lab environment similar to one of the simulated worlds. 
{\color{black}One of the researchers controlled the human-operated hider robot, which had access only to the front-facing camera feed and map of the environment, while the autonomous seeker robot made decisions autonomously using either one of our baselines or our approach.}

\subsubsection{Environments}
We construct four environments in simulation, as shown in the top half of Fig. \ref{fig:environment_CDP}. The first environment is a \texttt{small-dense-room} with several objects densely placed in a small room. The second environment is a \texttt{medium-dense-room} with objects spread across the map. The third is a \texttt{large-sparse-room}, which is sparse and has the same objects as \texttt{medium-dense-room} but wider open spaces. The last one is the \texttt{crossroad} environment, with four hallways emanating from a single crossroad and distinct objects or environmental geometries at the end of the hallways. The laboratory environment for physical experiments is similar to the \texttt{small-dense-room}.

\subsubsection{Policy Bank}
We model the various strategies of the human-controlled hider using behavior trees \cite{colledanchise2018behavior, lew2023shutter}. Each behavior tree takes the hider to various hiding locations within a given map, determined by the presence of particular objects in those locations or locations with different environmental occlusions. If the hider were to spot the seeker while navigating to a pre-defined location, the behavior tree would enable the hider to perform a fixed evasive maneuver to escape the seeker's field of view. Each behavior tree was executed multiple times with different starting locations on the map for both the hider and the seeker to collect a dataset of states and actions for each behavior of the hider. To construct the hider's policy bank, each policy was trained using the state-action trajectories in the corresponding dataset with a Long Short-Term Memory network \cite{hochreiter1997long} and behavioral cloning \cite{abbeel2004apprenticeship}. 

\subsection{Implementation Details}

The map of the shared environment was discretized into an $n \times n$ grid, and each map cell was represented as a node in an undirected graph. Each node in the environment graph would contain information about the center position of the cell, the objects present in the cell, and a visibility score of that cell, computed from every other cell in the map using ray casting in a raster grid. The state space $S_H$ of the hider agent was 6-dimensional, containing $(x_{h}, y_{h}, \theta_{h}, \hat{x}_{s}, \hat{y}_{s}, \hat{\theta}_{s})$, where $x_{h}, y_{h}$ denoted the position of the hider, $\theta_{h}$ is their heading, $\hat{x}_{s}, \hat{y}_{s}$ is the estimated position of the seeker by the hider, and $\theta_{s}$ was the estimated heading of the seeker by the hider. To obtain the perceived position and heading of the hider, we first fine-tuned the Yolact \cite{bolya2019yolact} real-time object detection model to predict a bounding box and a heading estimate for a given robot when it is visible in the first-person view of another robot. Then, based on the robot's known physical size, the predicted bounding box in the image, and the camera parameters, we use a pinhole camera model to estimate the relative node position of the hider w.r.t. the seeker. The action space of the seeker consists of transitions to each adjacent cell from an inhabited cell, where diagonal transitions are weighted proportionately in path computations.

To compute Critical Decision Points, we predict the actions the hider would take per a given policy for every node in an environment, $\pi_i \in \mathcal{P}$, assuming that the hider cannot see the seeker. This results in a map of Critical Decision Points, as shown in the bottom half of Fig. \ref{fig:environment_CDP}. The map shows that the environments and the policy bank created for the hide-and-seek task have afforded the existence of sparsely located Critical Decision Points. 

The seeker jointly optimizes to drive the hider toward the closest Critical Decision Point (with $\theta$ $\geq$ 4 in Eq. \ref{eq: cdp}) in a fixed radius around the hider's current position while minimizing its distance to the hider ($C_{task}$ in eq. \ref{eq:rhp_cost}). Additionally, the seeker optimizes for staying out of the hider's field of view when it approaches a Critical Decision Point using RHP.


Whenever the hider is visible, the seeker expands a tree of all possible future actions it could take in response to a hider following a given policy up to a certain horizon ($\mathcal D_H = 5$). 
Concurrently, the seeker performs a Bayesian update on the belief of the hider's policies, given observations of the hider, to compute the likelihood of each policy being executed from the policy bank. Finally, this belief is used to evaluate the cost function for each branch in the tree. Fig. \ref{fig:mpc} shows an example of tree expansion. Note that if the hider has been seen in the past 4 ($\mathcal D_{H-1}$) timesteps and is not currently visible, the seeker uses the stored action queue from the RHP tree in anticipation of spotting the hider in the future if its belief over the policy currently being followed by the hider is correct. Once no more stored actions are in the queue, the autonomous seeker reverts to a default exploration strategy where the seeker chooses an object on the map according to the policy it is following and navigates toward it. 

\subsection{Comparisons}

We evaluate our method in simulation and demonstrate our approach in a real-world environment. 

\subsubsection{Simulation}
 We measure the effectiveness of our approach by comparing the time it takes the autonomous seeker to catch the human-controlled hider against two baselines:

\begin{enumerate}[leftmargin=*]
    \item \textbf{Baseline 1: } The seeker randomly explores the environment and performs a Bayesian update of the belief over the hider's policies whenever it is visible (per eq. \ref{eq:belief_final}). 
    \item \textbf{Baseline 2: } The seeker uses RHP to optimize for minimizing its distance between itself and the estimated position of the hider whenever it is visible while computing the Bayesian update (per eq. \ref{eq:belief_final}). 
    \item \textbf{Our Method: } The seeker selects actions by expanding the RHP tree to drive the hider towards the closest Critical Decision Point per eq. (\ref{eq:rhp_cost}), and estimates their strategy using the Bayesian update (per eq. \ref{eq:belief_final}). 
\end{enumerate}

{\color{black}For our simulated experiments, we simulate a human-operated hider whose underlying route is randomly selected from the policy bank for each interaction episode.}
Depending on the environment, the hider's policy bank contains between 6 and 8 policies.  
We run each experiment for 20 episodes or trials, each of which ends when the seeker lands on the same node as the hider (indicating it has caught the hider) or after 100 time steps. To compare our method's performance with the baselines, we compute the average number of time steps needed for the seeker to catch the hider in each environment, aggregated along all ground truth policies from the policy bank, followed by the human hider for all trials. {\color{black}In partially observable environments like our hide-and-seek environment, the seeker updates beliefs intermittently only when the hider is visible, making step-by-step goal accuracy metrics unrepresentative due to line-of-sight gaps. Therefore, we utilize the time required to catch the hider as a proxy metric.}

\subsubsection{Real World}
 We demonstrate the applicability of our method in the real-world environment (shown in Fig. \ref{fig:realrobot}). We used Simultaneous Localization and Mapping
 \footnote{\href{https://wiki.ros.org/gmapping}{https://wiki.ros.org/gmapping}}
 to generate a map of the physical environment. Like the simulation environments, we discretized the map into an $n \times n$ grid 
to create an undirected graph with the states as nodes and valid actions from the states as edges. Depending on the size of the environment, $n$ varied between 7 and 15.  Then, we developed the policy bank of the hider by collecting trajectories of the robot moving between certain pre-defined locations of the map 
using the behavior trees and generating policies with behavioral cloning. We computed the Critical Decision Points for the real-world environment by computing the divergence of actions at every cell in the discretized grid. 
One of the researchers controlled the hider robot through velocity commands using a PlayStation controller while following a given policy from the policy bank.  The seeker robot selected actions optimized for driving the opponent toward the Critical Decision Points following our RHP-based approach.

\begin{figure}
    \centering
    \includegraphics[width=\linewidth]{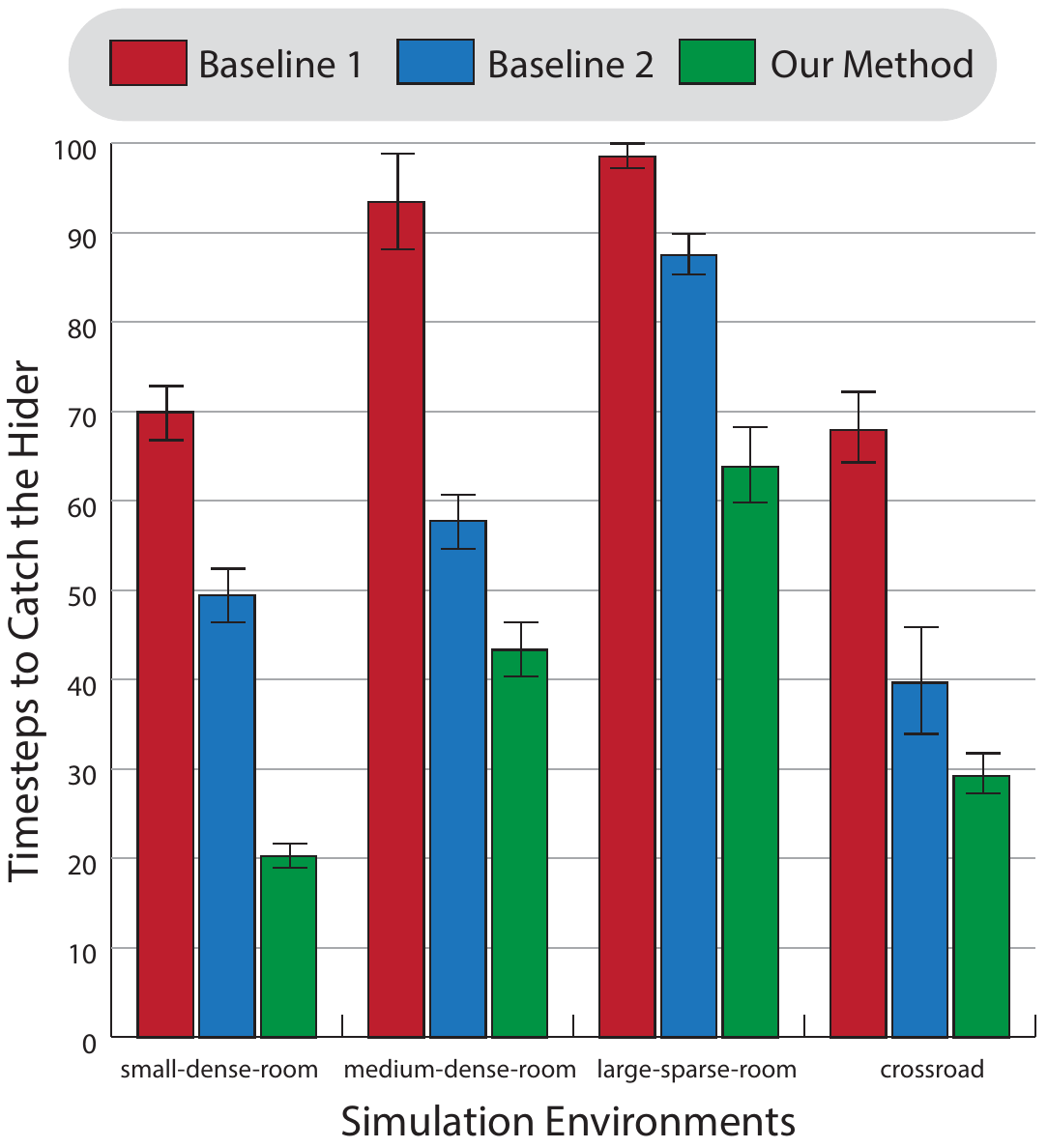}
    \caption{\textbf{Results of Competitive Evaluation: }Mean and Standard error of timesteps taken by the seeker to catch the hider for each experiment aggregated across all hider policies for a given environment and number of trials (lower is better)}
    \label{fig:catching_stats}
\end{figure}

\subsection{Results}

\subsubsection{Simulation}
Fig. \ref{fig:catching_stats} shows that our method outperformed the baselines across all four simulation environments. When the seeker influenced the hider toward Critical Decision Points, it caught the hider faster than when it randomly explored the map (Baseline 1) or minimized its distance to the hider whenever the hider was visible (Baseline 2).

Qualitatively, we observed that with Baseline 1, when the seeker randomly traversed the map and updated its belief of the hider's policy with each sighting, the seeker could not accurately estimate the hider's strategy. In Baseline 2, when the seeker used RHP to minimize its distance to the perceived position of the hider, it frequently entered the hider's field of view, making the hider trigger a fixed evasive maneuver. The seeker's estimate of the hider's strategy then became inaccurate. With our method, the seeker entered the hider's field of view to trigger an evasion and then exited it to observe the hider at the Critical Decision Point continue to its original goal. It then selected actions to intercept the hider quickly once it learned the hider's strategy.

\subsubsection{Real World}
We demonstrated our method on the real robot across five trials each for six policies, as illustrated in Fig. \ref{fig:realrobot}. A researcher controlled the hider robot, with access to only the map of the environment and the first-person view from the robot's camera. Our method took the seeker 27 $\pm$ 3 timesteps to catch the hider. Baselines 1 and 2 took the seeker 39 $\pm$ 2 timesteps and 24 $\pm$ 1 timesteps longer than our approach to catch the hider, respectively. The supplementary video shows an example trial of the real-world demonstration. 

\subsection{Comparative Analysis of Performance of CDPs in the Competitive Task}

In this work, we developed a novel method for a robot to identify a human's strategy with limited observability of human actions. Assuming the robot has access to reasonable policies a human might follow, the robot examines the actions these policies predict at all states in the environment. Then, by observing humans at the states where most policies diverge (which are CDPs), the robot can more quickly predict which of the strategies the human is using in the environment. However, the number and spatial distribution of CDPs depend on both the layout of the shared environment and the range of plausible human strategies under consideration. In some environments, CDPs are abundant, and passive observation suffices. In others, they are rare, and the robot must actively optimize its behavior to increase the chance of observing the human at a CDP.

We demonstrated our approach in a game of hide-and-seek in simulation and the real world, assuming that the human's strategy remained unchanged throughout a single game. Our results suggest that influencing a human-controlled hider toward Critical Decision Points can help an autonomous robot infer the human's strategy more quickly in the game.

\section{Discussion}


{\color{black}
\subsection{Applicability Across Diverse Task Settings}
This work evaluates our method in two domains that represent opposite ends of several key axes. The collaborative cooking domain is fully observable, features a large state space, and involves a cooperative human partner who shares the robot’s objective. In contrast, the competitive navigation domain is partially observable, has a relatively small state space, and models an adversarial human whose goal is intentionally hidden from the robot. Despite these differences, our approach of leveraging policy divergence to identify CDPs and influence humans toward those CDPs proves effective for identifying the human's goal in both settings. This highlights the flexibility of our framework, which does not assume a fixed interaction mode, observability level, or environment size. As such, it may be adapted to domains that combine properties along these axes, for example, semi-collaborative settings, mixed observability, or medium-sized domains, provided task-specific models and policy banks can be defined.}

{\color{black}
\subsection{Active Goal Inference Through CDPs}

Our results suggest that fast goal inference depends not only on how the robot updates its belief, but also on whether the robot can guide the interaction towards situations where informative evidence can become available. A purely passive observer must wait until the human naturally reaches a state where their actions distinguish between possible goals. In contrast, our approach allows the robot to select actions that guide the interaction toward such states earlier. The role of CDPs is therefore to identify where the human’s next action is expected to be most informative, while the RHP objective ensures that the robot still accounts for task progress.

This perspective helps explain why the same framework applies across the two scenarios. In the collaborative cooking task, the robot can take supportive actions that remain compatible with multiple possible meals while encouraging the human to take a more goal-specific action later in the interaction. In the competitive hide-and-seek task, the seeker can use the structure of the environment and the hider’s possible strategies to increase the chance of observing behavior that distinguishes between those strategies. Thus, the contribution of CDPs is not only to improve goal prediction after observing human behavior, but to enable the robot to act in ways that make future human behavior more diagnostic.
}

\subsection{Limitations and Future Work}


We make some limiting assumptions about our experimental setup that may not apply in the real world. 
We demonstrate our approach in a deterministic environment. We assume that humans always act rationally, make no mistakes, and do not change the task during the interaction. We also assume that the interaction is turn-based, with neither agent able to take a ``wait" action, and actions once taken by either agent cannot be repeated in the collaborative case. Our approach also requires an explicit policy bank, which may not be readily available in many real-world scenarios. It assumes that the task can be represented in a discrete space. 
Future work will involve relaxing these assumptions  
{\color{black}in three main directions: handling goal changes during interaction, extending CDP computation and planning to continuous or hybrid state-action spaces, and reducing reliance on hand-specified policy banks by learning, retrieving, or generating plausible goal-conditioned policies from demonstrations, language, or vision-language models. Additionally, future work must evaluate the framework's robustness to upstream errors in the perception of the state space and human behavior, as well as failures in robot behaviors.}

{\color{black}Another important direction for future work is combining action-based goal inference with language-based interaction similar to \cite{ghose2026open}. In this paper, we intentionally isolate the question of how much a robot can infer from actions alone. In real-world collaborative settings, however, a robot will likely need to combine actions with language, since verbal instructions can provide useful high-level information while still leaving many grounded task details unspecified. For example, a person may state a broad goal, while the robot uses subsequent actions to infer the particular way the person intends to carry it out. In this sense, action-only inference should be a complementary capability for verbal interactions to reduce the amount of step-by-step specification or repeated clarification needed from the human. Future systems could use language to initialize or update the robot’s belief over possible goals, while using CDPs to identify when actions in the shared environment are most informative for disambiguating those goals. More concretely, large language or vision-language models could be used to support specific components of our framework, such as proposing candidate goals from a high-level instruction, mapping language or visual observations into the discrete state and action representation used by the planner, or generating plausible policies for the policy bank. The CDP framework could then operate over these grounded representations to decide which robot action would move the interaction toward states where the human’s next action is most informative for disambiguating between goals.}

\bibliography{sn-bibliography}

\end{document}